\documentclass[manuscript,screen,authorversion,nonacm]{acmart}

\AtBeginDocument{%
  }

\setcopyright{acmlicensed}
\copyrightyear{2018}
\acmYear{2018}
\acmDOI{XXXXXXX.XXXXXXX}

\renewcommand\footnotetextcopyrightpermission[1]{} 
\newcommand{\revised}[1]{\textcolor[RGB]{0,0,0}{#1}}
\newcommand{\rerevised}[1]{\textcolor[RGB]{0,0,0}{#1}} 

\usepackage{amsmath}

\DeclareMathOperator*{\argmin}{arg\,min}
\DeclareMathOperator{\NeuralNet}{NN}
\usepackage{graphicx}
\usepackage{prettyref}
\usepackage{algorithm}
\usepackage{algpseudocode}
\usepackage{subcaption}
\usepackage{booktabs}
\usepackage{etoolbox}
\usepackage{makecell}
\usepackage{multirow}
\usepackage{wrapfig}
\makeatletter
\patchcmd{\@makecaption}
  {\scshape}
  {}
  {}
  {}
\makeatother

\newcommand{\Rmnum}[1]{\uppercase\expandafter{\romannumeral #1}}

\usepackage{xcolor}

\newrefformat{fig}{Fig.~\ref{#1}}
\newrefformat{tab}{Table~\ref{#1}}
\newrefformat{alg}{Algorithm~\ref{#1}}
\newrefformat{eq}{Eqn.~\ref{#1}}
\begin{document}

\title{Internal State Estimation in Groups via Active Information Gathering}

\author{Xuebo Ji}
\affiliation{
  \institution{The University of Hong Kong}
  \country{Hong Kong}
}
\email{xbji@connect.hku.hk}

\author{Zherong Pan}
\affiliation{
  \institution{Tencent America}
  \country{USA}
}
\email{zherong.pan.usa@gmail.com}

\author{Xifeng Gao}
\affiliation{
  \institution{Tencent America}
  \country{USA}
}
\email{gxf.xisha@gmail.com}

\author{Lei Yang}
\affiliation{
  \institution{Centre for Transformative Garment Production}
  \institution{The University of Hong Kong}
  \country{Hong Kong}
}
\email{l.yang@transgp.hk}

\author{Xinxin Du}
\affiliation{
  \institution{Tsinghua University}
  \country{China}
}
\email{duxx@tsinghua.edu.cn}

\author{Kaiyun Li}
\affiliation{
  \institution{University of Jinan}
  \country{China}
}
\email{Sep_liky@ujn.edu.cn}

\author{Yongjin Liu}
\affiliation{
  \institution{Tsinghua University}
  \country{China}
}
\email{liuyongjin@tsinghua.edu.cn}

\author{Wenping Wang}
\affiliation{
  \institution{The University of Hong Kong}
  \country{Hong Kong}
}
\email{wenping@cs.hku.hk}

\author{Changhe Tu}
\affiliation{
  \institution{Shandong University}
  \country{China}
}
\email{chtu@sdu.edu.cn}

\author{Jia Pan}
\affiliation{
  \institution{The University of Hong Kong}
  \country{Hong Kong}
}
\email{jpan@cs.hku.hk}

\renewcommand{\shortauthors}{Ji et al.}

\begin{abstract}
\revised{Accurately estimating human internal states, such as personality traits or behavioral patterns, is critical for enhancing the effectiveness of human-robot interaction, particularly in group settings. These insights are key in applications ranging from social navigation to autism diagnosis. However, prior methods are limited by scalability and passive observation, making real-time estimation in complex, multi-human settings difficult. In this work, we propose a practical method for active human personality estimation in groups, with a focus on applications related to Autism Spectrum Disorder (ASD). Our method combines a personality-conditioned behavior model, based on the Eysenck 3-Factor theory, with an active robot information gathering policy that triggers human behaviors through a receding-horizon planner. The robot's belief about human personality is then updated via Bayesian inference. We demonstrate the effectiveness of our approach through simulations, user studies with typical adults, and preliminary experiments involving participants with ASD. Our results show that our method can scale to tens of humans and reduce personality prediction error by $29.2\%$ and uncertainty by $79.9\%$ in simulation. User studies with typical adults confirm the method’s ability to generalize across complex personality distributions. Additionally, we explore its application in autism-related scenarios, demonstrating that the method can identify the difference between neurotypical and autistic behavior, highlighting its potential for diagnosing ASD. The results suggest that our framework could serve as a foundation for future ASD-specific interventions.}
\end{abstract}

\ccsdesc[500]{Computing methodologies~Artificial intelligence}
\ccsdesc[500]{Computing methodologies~Active learning settings}
\ccsdesc[100]{Computing methodologies~Modeling and simulation}
\ccsdesc[300]{Applied computing~Psychology}
\ccsdesc[300]{Computer systems organization~Robotics}

\keywords{Internal State Estimation, Personality Estimation, Behavior Prediction, Autism Spectrum Disorder, Robot-Crowd Interaction, Active Learning}


\maketitle

\section{Introduction}
\revised{Human internal state estimation has been an active interdisciplinary research area in the field of human-robot interaction over the past decades~\cite{7139219, bera2017aggressive, shen2019inferring, information_gathering}. Human behavior is generally influenced by internal states, which are not directly accessible to a robot during interaction~\cite{information_gathering}. For example, a driver's driving style --- whether aggressive or defensive --- can significantly impact their driving behavior in interactive traffic scenarios. Similarly, personality traits can result in a wide range of variations in human's navigation behavior~\cite{bera2017aggressive}, while children with autism can exhibit restricted, repetitive patterns of behavior in social interactions~\cite{regier2013dsm}. Accurately predicting these internal states is essential for the robot to obtain more information and sustain an effective interaction process. On one hand, valuable information can be extracted through behavioral data analysis, such as a robot identifying signs of autism based on observed actions. On the other hand, the estimation of the human state can enhance human-robot interaction, enabling the robot to achieve its objectives more efficiently and precisely. For example, with an accurate prediction of an autistic child's behavior, the robot could devise more effective intervention strategies. Recently, there has been a growing focus on methods where robots predict human behavior and employ the ``coupled'' method that incorporates a human model in the planning for joint robot-human actions~\cite{7139219,bera2018socially,sadigh2016planning}.}

\revised{However, the automatic prediction of human internal states during interactions presents significant challenges. The robot must perform multiple layers of logical inference, specifically to predict human responses following its actions and evaluate whether these responses facilitate state estimation and interaction. In addition, interactions commonly occur in group settings in real-world scenarios, such as driving scenarios, social navigation, and group activities for children with autism in special education schools. The complex, nested interactions between multiple individuals in these group contexts pose significant challenges for accurate state estimation. In this work, the general goal is to enable the robot to actively estimate human internal states in groups.}

A row of theoretical frameworks has been proposed to estimate human states and behaviors. The partially observable Markov decision process (POMDP) lays the theoretical foundation for solving such problems~\cite{kaelbling1998planning}. POMDP models the robot-human interaction as a joint dynamic system conditioned on the human internal states, aka intentions or behavioral patterns. It then jointly updates the robot behaviors and estimates the internal states via Bayesian rules. This framework, however, \rerevised{is limited in its scalability}. The human internal states have to be estimated in time in order for them to be effectively exploited~\cite{7139219}, but directly solving POMDP is computationally expensive. \revised{In group settings, the robot simultaneously interacts with multiple people, where estimating the state for each individual via POMDP is intractable.} \rerevised{To improve the computational efficiency of POMDP, several approximations have been proposed for similar problem formulations~\cite{javdani2015shared, fern2014decision}. However, these methods only update the human internal states from \textit{``passive'' observations}~\cite{information_gathering}, i.e., it cannot explore human behaviors that are important for interaction but are not exhibited without external triggers.} There have been prior works that partially address the aforementioned problems, but a complete solution is yet to be presented.

In order to improve the efficacy of human internal state estimation, prior works have combined multi-modal information including audio~\cite{gu2017speech}, vision~\cite{li2020deep, bera2017aggressive}, gesture~\cite{lohse2014robot}, and unintentional human behaviors~\cite{mutlu2009nonverbal}. Modern deep-learning approaches~\cite{alahi2016social} can scale up to detect several human or vehicle trajectories at the same time, but they still rely on passive observations. One pioneering work on ``active'' information gathering is presented in~\cite{information_gathering}. It models the human and robot as a joint differential dynamic system and computes a robot trajectory that maximizes the information gain of human driving style via model predictive control (MPC). However, this method requires solving an expensive optimization problem, where both human's and robot's actions are optimized in a nested manner to minimize different objectives. Even for a single human, solving such a problem in real-time can be a major computational burden. Several follow-up works extend the idea of~\citet{information_gathering} to the application of grasping~\cite{bestick2017implicitly} and other cobot applications~\cite{nikolaidis2018planning,che2020efficient}, but they all suffer from limited scalability due to involving POMDP or independent optimization for each human.

\rerevised{To address these issues, we proposed a novel framework for estimating human internal state in group settings actively and efficiently. Our key insight is to employ active information gathering to acquire crucial information for more effective internal state estimation, while leveraging system-level inference to ensure the maintenance of both estimation quality and processing speed. We emphasis the importance of the concept ``active gathering in an interactive context'' for efficient internal state estimation, and additionally consider the potential applications in diagnosing ASD. This work is conducted as a prospective study for assessing the technical feasibility of our insight, where we select \textit{personality} as the internal state and \textit{navigation behavior} as the action space to study the \textit{scalable human personality estimation} problem. A series of evaluations are performed in simulation and on typical human groups. Although forward-looking, we conducted additional evaluations on the \textit{autistic} groups to explore the connections between the proposed method and ASD diagnosis. Specifically, our \textbf{main contributions} are twofold:}

\revised{\noindent{\textbf{Human Personality Estimation:}} We study the problem of \rerevised{personality estimation} in typical human groups.} As numerous works have linked personality models with heterogeneous crowd behaviors which have obtained reliable results~\cite{allbeck2008creating, salvit2011toward, crowd_sim_psn_trait_theory, bera2017aggressive}, we follow the personality trait theory~\cite{Pen_model} as~\citet{crowd_sim_psn_trait_theory}, which stresses that extensive changes in human behavior is mainly caused by personality traits. Specifically, we use the established Eysenck 3-Factor personality model~\cite{Pen_model} based on a 3-D behavior descriptor (PEN): Psychoticism, Extraversion, and Neuroticism. We link human behavior in groups with these three factors and focus on the observed navigation strategy. As a main point of departure from prior works, we propose an efficient and scalable framework to actively estimate personalities for tens of humans. Firstly, we train \textit{a personality-conditioned human behavior network} via the trait theory~\cite{crowd_sim_psn_trait_theory}, which could efficiently predict collision-free trajectories. As a result, the trajectories of multiple humans can be predicted in a generative and differentiable manner without solving additional decision-making problems. Second, we propose a receding-horizon planner to compute robot trajectories that actively interact with crowds to trigger human's behavior. Finally, the personalities of multiple observed humans are updated according to observed behaviors via Bayesian rules. \revised{One core aspect of our method is enabling robots to perform active, interactive actions to collect more effective information. However, balancing this active information gathering with \textit{human comfort} during interaction is essential, posing a trade-off. To address this, we reduce the aggressiveness of the robot's behavior by fine-tuning parameters and applying additional optimization, ensuring human comfort is prioritized throughout the interaction process.}

\revised{\noindent{\textbf{Evaluation on Both Neurotypical and Autistic Groups:}} As a preliminary work to explore the technical feasibility of applying the proposed method in autism-related applications, we conduct a series of evaluations to validate the effectiveness of our framework and further examine its impact when applied to autistic participants. Specifically:
\begin{itemize}
    \item We evaluate our method on a row of benchmarks in \textit{simulation}. We show that our method can scale to tens of humans with unknown personalities in real-time, and the active gathering is much more efficient than passive gathering which lowers the personality prediction error and uncertainty about personality by $29.2\%$ and $79.9\%$ on average, respectively.
    \item Two user studies are conducted on \textit{typical adults}, which show that our active method can be generalized to complex human personality distributions and can elicit human behaviors which are conducive to personality estimation.
    \item We implemented the proposed method in the form of a PC game and applied it to \textit{participants with autism}. Our results demonstrate that the method can identify the difference between autistic and neurotypical participants based on the predicted internal states, which may be linked to the restricted and repetitive behavior patterns commonly observed in autism.
\end{itemize}}

\revised{Overall, this work takes a step toward active internal state estimation in group settings, focusing on scalable personality estimation as a preliminary study for the diagnosis and treatment of Autism Spectrum Disorder (ASD). \rerevised{Although ASD-specific internal state and action space were not used to develop the model in this work, we still identified meaningful correlations between the proposed framework and ASD-related applications.} These findings lay the groundwork for our future research, where we aim to use behavior data collected from ASD groups to refine the model and apply our framework to actively predict internal states for autistic children, exploring its potential as an intervention strategy.}
\section{Related Work}
Our method is closely related to recent progresses in human behavior prediction, human personality modeling, \revised{multi-agent navigation, and human-robot interaction in autism}. We briefly review prior works on each topic.

\subsection{Human Behavior Prediction}
Human behaviors take the form of navigation trajectories, gestures, languages, etc., and early works like social Long Short-Term Memory (LSTM)~\cite{alahi2016social} focus on detecting and predicting these external behaviors. The human behavior predictor can also be combined with robot planners~\cite{bandyopadhyay2013intention} to improve the efficacy of human-robot collaboration. However, without the explicit modeling of internal states, pure external behavior prediction would require more data for training and is weaker in generalization. This limit can be resolved via POMDP~\cite{7139219} where the robot trajectory computing and the human internal state estimating are conducted simultaneously.

All the aforementioned methods are passive predictions, where the predictor observes humans without interacting with them. The seminal work of~\citet{sadigh2016planning} first notices the possibility for the robot to influence human behaviors and their follow-up work~\cite{information_gathering} computes robot influencing trajectories to actively probe and detect human's driving style. However, their method needs to solve a costly nested optimization to compute the active probing trajectory for a single human. Follow-up works~\cite{bestick2017implicitly,nikolaidis2017human} allow the robot to influence humans not only for probing but also to better accomplish the task. However, the robot only interacts with a single human in these applications. Their poor scalability limits their application in group settings where the robot interacts with multiple humans simultaneously.

\subsection{Human Personality Modeling}
Personality is one of the most decisive factors in human behaviors~\cite{crowd_sim_psn_trait_theory}. Human personalities have been summarized by various dimensional models, which have been further incorporated into agent simulation systems. The dimensional models assume that the personality is affected by several factors. For example, the OCEAN model~\cite{wiggins1996five} identifies five factors and~\citet{allbeck2008creating} and~\citet{durupinar2009ocean} relate them with low-level simulation parameters via a generative probabilistic model. The Myers-Briggs type indicator model~\cite{myers2010gifts} identifies four factors and~\citet{salvit2011toward} implements a swarm of agents covering a subset of the Myers-Briggs model. The personality trait theory~\cite{eysenk1985personality} identifies three major factors and~\citet{crowd_sim_psn_trait_theory} link them with the low-level simulator parameters of the (Reciprocal) Velocity Obstacles (RVO) algorithm~\cite{ORCA} using regressive analysis. Our method is based on~\cite{crowd_sim_psn_trait_theory} and we train a personality-conditioned navigation policy to predict the behavior of agents with different personalities.  

\subsection{Multi-Agent Navigation}
The personality estimation in this work is related to social navigation. The classic centralized navigation algorithms~\cite{bender1998power,sharon2015conflict} assume an omniscient node coordinates and plans motions for a swarm of robots, while they suffer from heavy communication burden and vulnerability to system error. Some rule-based decentralized microscopic approaches, e.g. the social force model~\cite{helbing1995social} and velocity obstacles (RVO)~\cite{RVO} make humans navigate around by utilizing local information in a small vicinity, while several follow-up works utilize local collaboration to achieve better navigation efficacy or social norm compliance~\cite{godoy2016implicit, he2016dynamic}. However, it is difficult to translate task-level objectives, such as time-optimality~\cite{fan2020distributed} and social-awareness~\cite{chen2017socially}, into analytic rules. In view of this, learning-based algorithms like inverse reinforcement learning (IRL)~\cite{kuderer2012feature} and reinforcement learning (RL)~\cite{fan2020distributed, 9740403} propose to learn low-level rules from high-level specifications. However, \revised{a common drawback of the above-mentioned techniques is that} their behaviors are not personalized for each human. We resolve this problem by training a personality-conditioned navigation policy.

\subsection{Human-Robot Interaction in Autism}
\revised{Traditional methods for diagnosing and treating autism mainly rely on behavioral therapy, which is time-consuming~\cite{leaf2022concerns}. Recently, artificial intelligence has increasingly been applied in autism diagnosis, screening, and treatment~\cite{anagnostopoulou2020artificial}, utilizing multimodal data like electroencephalography (EEG)~\cite{bosl2018eeg}, eye-tracking~\cite{kim2023development}, and facial and hand movement~\cite{del2017computer, li2017applying}. Compared to solely relying on observation, interactive robots can better capture the attention of children with autism, improving data collection efficiency and offering more diverse intervention options~\cite{saleh2021robot}. Robot-based interventions are used to enhance the autistic children's social skills~\cite{suzuki2016robot}, emotional understanding~\cite{bevill2016multisensory}, and imitation abilities~\cite{chevalier2017sensory}. While these approaches have shown some success, they mainly rely on manually designed interaction scenarios, which lack explicit modeling of a child’s internal states. As a result, the interactions are limited in intelligence and adaptability, making it difficult to meet the diverse needs of individuals with autism. Although some studies use interaction data to model and estimate states~\cite{marinoiu20183d}, data collection remains  “passive”, which becomes challenging when children display low levels of expression or have limited behavioral abilities. In this work, we study active personality estimation as a preliminary approach for autism diagnosis and treatment. Our method shows the potential of using active information gathering and dynamic internal state estimation during interactions, paving the way for more flexible and effective robot-based interventions for autism.}

\section{Problem Formulation and Assumptions}
We study the problem of human personality estimation in groups using active information gathering. We assume that there is a single robot $R$ and $N$ humans $H_{1...N}$ with personality descriptors $\phi_{1...N}$ navigating in a shared space. Both the robot and humans can observe the physical state of others within a certain range. According to~\cite{crowd_sim_psn_trait_theory}, we further assume that the human $H_i$'s navigation pattern is affected by the personality $\phi_i$, which is unknown to the robot. Our goal is to estimate the personality of each person through the robot's active probe.

\paragraph{Joint Dynamic System}
We represent the robot and humans as a joint dynamic system with state $s=\{s_R,s_{H_1},\cdots,s_{H_N}\}$. At each time step, the motion of the robot and all humans is characterized by $f$:
\begin{align}
s^{t+1} = f(s^t,\mu_R^t,\mu_{H_{1...N}}^t), \notag
\end{align}
where $\mu_R^t$ and $\mu_{H_{1...N}}^t$ are continuous control inputs for the robot and humans, respectively, and the dynamic system $f$ encodes physical rules determining the complex interaction between the robot and humans, such as the collision avoidance. Following the personality trait theory~\cite{crowd_sim_psn_trait_theory, bera2017aggressive}, we further assume human's personality $\phi_i$ is \revised{the only latent factor} affecting its behavior, which is also intransient and independent. Indeed, during the short period of robot-human interaction, the human emotion and goal-biased intention are relatively stable and can be assumed indecisive.

\paragraph{Human Planner}
We assume that each human $H_i$ is a local planner that determines its behavior via a personality-conditioned function:
\begin{align}
\mu_{H_i}^t = F(s_{H_i}^t, o_{H_i}^t, \phi_i), \notag
\end{align}
where $s_{H_i}^t$ represents human $H_i$'s physical state (e.g., current position $p_{H_i}$, velocity $\dot{p}_{H_i}$, goal) and $o_{H_i}^t$ is $H_i$'s observation at the time instance $t$ (e.g., states and historical trajectories of humans or robot in $H_i$'s sensing range $d_H$). It is worth noting that since humans and robot have different goals, following previous work ~\cite{information_gathering}, we assume that humans will not make decisions to actively influence the robot.

\paragraph{Robot Planner}
The main goal of our robot is estimating human's personality. We assume that the robot does not know the number of humans in the environment and has only a limited sensing range $d_R$. To simplify the problem setting and focus on active personality inference, we further assume the robot has perfect sensing within the sensing range. \rerevised{We set up the robot to keep track of a belief $b(\phi_i)$ for each human it has observed and continually refine the belief as it collects more observations about this human.} Notably, it does not mean we only consider each human’s behavior individually: As humans are interactive individuals that always consider the surrounding environment with collision-avoidance constraints, different humans are interacting with each other and reacting based on their observations. Our method does take such interactions into account. Compared with observing and predicting passively in prior works \cite{li2020deep, bera2017aggressive}, we make the robot generate behaviors intelligently which could have an active interaction with the crowds for drawing better inferences. In addition, another challenge of our problem lies in quickly and tractably estimating and updating $b(\phi_i)$ for dozens of humans, rather than dealing with only a single human in prior work~\cite{information_gathering}.
\section{Active Human Personality Estimation}

\begin{figure*}[t]
\centering
\includegraphics[width=0.98\linewidth, trim=58 70 20 78,clip]{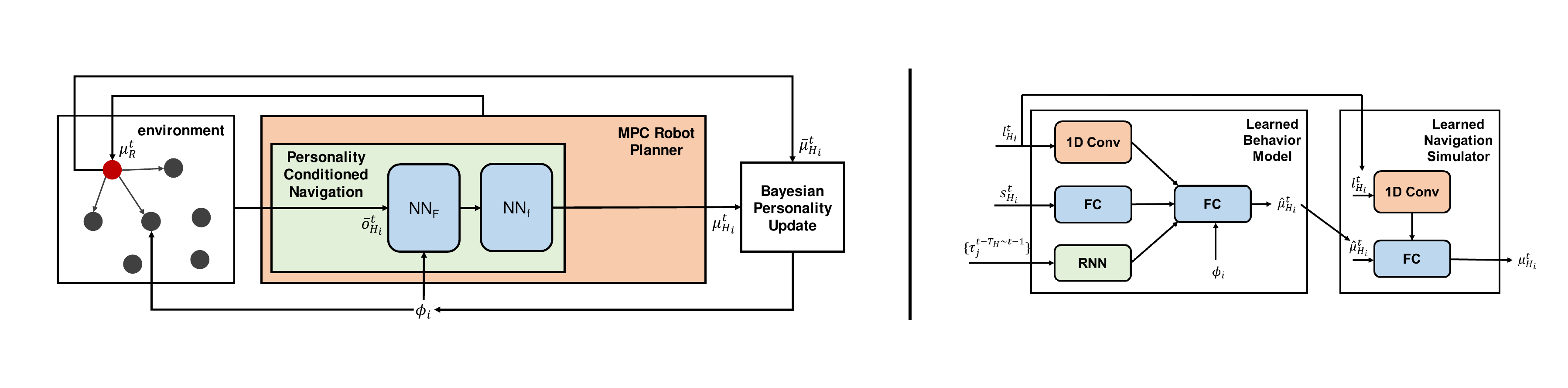}
\caption{\label{fig:pipeline_networks} The pipeline of our system (left) and the neural network structure of our human behavior model (right). During each timestep, the robot observes a short history of neighboring human trajectories, from which beliefs over human personalities are updated using Bayesian rules. Finally, our robot planner optimizes the robot control policy to actively probe human personalities.} 
\end{figure*}

In this section, we first describe the details of the dynamic system $f$ and local planner models used in our framework, and then introduce our personality estimation scheme. The full pipeline of our method is outlined in~\prettyref{fig:pipeline_networks}.

\subsection{\label{sec:human_model_sec}Learning-based Personality-conditioned Navigation}
Our method is based on the key observation drawn in prior research~\cite{pervin2003science}: human behavior is influenced by several internal factors. Generally speaking, human's overall behavior and moving patterns will be biased by their underlying personalities, even when they are pursuing the same task. Recent works~\cite{Pen_model} and~\cite{crowd_sim_psn_trait_theory} showed that for the particular task of navigation, the low-level simulator parameters of the RVO algorithm~\cite{ORCA} and the Eysenck 3-factor personality model (PEN model)~\cite{Pen_model} have a strong correlation. Different from discrete internal state features~\cite{information_gathering}, PEN model uses a 3D continuous indicator to model the three factors: Psychoticism, Extraversion, and Neuroticism. \citet{crowd_sim_psn_trait_theory} derive a mapping between the personality factors and RVO parameters using data-driven linear regression which involves a wide range of participants, leading to generative personality-conditioned human behavior. However, their linear regression involves a pseudoinverse, i.e., the number of RVO parameters is larger than that of personality factors. As a result, using insufficient observation to estimate these parameters can lead to undetermined behaviors. \revised{Further, the RVO algorithm is non-differentiable, which limits the potential efficacy when applying model-predictive controls to estimate the trajectories for humans.}

\paragraph{Human Local Planner}
To resolve the above issues, we follow the idea of the personality-conditioned RVO behavior and propose a novel learning-based human local planner. As illustrated in~\prettyref{fig:pipeline_networks} (right), we use a local feed-forward neural network to parameterize the human local planner $F$, which maps human $H_i$'s physical state $s_{H_i}^t$ and its observation:
\begin{align}
\label{eq:human_obs}
o_{H_i}^t=\left\{l_{H_i}^t, \{\tau_{j}^{t-T_H \sim t-1} \mid \|p_{H_i}^t-p_{j}^t\|\leq d_H, j \in H_{-i}\cup \{R\}\}\right\},
\end{align}
to the intermediate control action $\hat{\mu}_{H_i}^t$. Here, $H_i$'s observation consists of its local sensing measurement $l_{H_i}^t$ and $\tau_j$, which is a short (within time horizon $T_H$) historical trajectory of other humans $H_{-i}$ or robot $R$ within $H_i$'s sensing range $d_H$. Following the setting in previous crowd simulation works~\cite{lee2018crowd}, the local sensing measurement $l_{H_i}^t$
is implemented in a way similar to LiDAR, where $m$ rays within $180$ degrees are used to simulate the perception from the front of a person. Finally, our network is also conditioned on $H_i$'s personality $\phi_i$. In summary, our local planner can be formulated as the following function:
\begin{align}
\hat{\mu}_{H_i}^t = \NeuralNet_{F}(s_{H_i}^t, o_{H_i}^t, \phi_i). \notag
\end{align}
The three inputs to $\NeuralNet_{F}$ has different modalities, for which we use different feature encoders. The current physical state $s_{H_i}^t$ is encoded by additional fully connected layers. To encode the local sensing input $l_{H_i}^t$, we use an 1D convolutional network. To encode the neighbor historical trajectory data $\tau_{j}^{t-T_H \sim t-1}$, we use a recurrent neural network (RNN). All the encoded features along with the personality $\phi_i$ are then fed into the feed-forward network to yield the intermediate action $\hat{\mu}_{H_i}^t$.

\paragraph{Learned Navigation Simulator}
The above local planner is for generating control actions for humans. However, simply executing such actions would lead to violations to the collision-free constraints among agents or against static obstacles. To ensure collision avoidance, we follow~\cite{fan2020distributed} and train another neural network to learn the joint robot-human dynamic system function $f$ in a decentralized manner. We introduce another feed-forward network $\NeuralNet_f$ that filters the intermediate human navigation action $\hat{\mu}_{H_i}^t$ to yield the ultimate collision avoidance action $\mu_{H_i}^t$. Our feed-forward network is also conditioned on human $H_i$'s local sensing data:
\begin{align}
\mu_{H_i}^t = \NeuralNet_{f}(l_{H_i}^t, \hat{\mu}_{H_i}^t). \notag
\end{align}
$H_i$'s final control action $\mu_{H_i}^t$ simply takes the form of a desired velocity and we update the human's position by simply translating it by $\mu_{H_i}^t\Delta t$, where $\Delta t$ is the timestep size.

\paragraph{Model \& Simulator Training}
Unlike~\cite{fan2020distributed} which uses RL, we train both $\NeuralNet_{f}$ and $\NeuralNet_{F}$ using supervised learning. We sample a large dataset of trajectories from the groundtruth personality-conditioned RVO simulator using randomized goal positions and personality parameters. Our loss function is the $l_2$-difference between the groundtruth action and $\mu_{H_i}^t$. Our key idea is to equip the robot with the learned personality-conditioned human local planner as a prior. This method significantly reduces the dimension of search space for the human personality. \revised{In addition, the navigation policy learned by neural networks offers a level of robustness and smoothness in interactions: the continuity of neural networks helps ensure that group behavior does not exhibit drastic fluctuations in response to sudden changes in a surrounding agent’s behavior \cite{long2017deep}. This capability enables effective responses to abrupt changes in user input during the interaction process.} Further, the learned simulator allows the robot to predict human behaviors in a generative manner, without nested optimization as done in~\cite{information_gathering}.

\subsection{\label{bayes_update}Personality Estimation}
As mentioned before, a human’s personality is assumed to be static and independent, thus the robot could infer each human's personality separately. We have our robot maintaining a Bayesian belief $b(\phi_i)$ over the personality of each observed human $H_i$. At timestep $t$, the robot can observe the behavior of humans within its sensing range $d_R$, i.e., all $H_i$ such that $\|p_{H_i}-p_R\|\leq d_R$. The robot's observation for human $H_i$ is $z_{H_i}^t=\{s_{H_i}^{t},\bar{o}_{H_i}^{t}\}$, which includes two parts: $s_{H_i}^{t}$ is the observed $H_i$'s current state; $\bar{o}_{H_i}^{t}$ is the robot's estimation of what $H_i$ should have locally observed. In particular, $\bar{o}_{H_i}^{t}$ is a subset of ${o}_{H_i}^{t}$ in \prettyref{eq:human_obs} where only $H_i$'s neighbors that can also be observed by the robot are taken into consideration: 
\begin{align}
\bar{o}_{H_i}^t=\{l_{H_i}^t,\bar{\tau}_{j}^{t-T_H \sim t-1}|\|p_{H_i}^t-p_{H_j}^t\|\leq d_H\land\|p_R^t-p_{H_j}^t\|\leq d_R\},
\end{align}
where $\bar{\tau}_{j}$ represents the historical trajectories of $H_i$'s neighbors who are visible to the robot, which is a subset of ${\tau}_{j}$ in \prettyref{eq:human_obs}.
 
Given the robot-observed human behavior $z_{H_i}^t$,  together with the human's actual action $\overline{\mu}_{H_i}^{t}$ observed by the robot, we propose to update robot's belief $b(\phi_i)$ using the Bayesian rule. In particular, the humans' actual action should follow the probabilistic model below:
\begin{align*}
&p(\bar{\mu}_{H_i}^t \mid z_{H_{i}}^t,\phi_i)=\frac{1}{Z}\exp\left(-\alpha\mathcal{L}\left(\bar{\mu}_{H_i}^t,\NeuralNet_f\left(l_{H_i}^t,\NeuralNet_F(s_{H_i}^t, \bar{o}_{H_i}^t, \phi_i)\right)\right)\right),
\end{align*}
where $Z$ is a normalizing factor, $\alpha$ is a sharpness-controlling parameter, and $\mathcal{L}(\cdot,\cdot)$ is the loss function, which is the cosine-similarity function in our case.  

Then, the Bayesian rule for updating the human personality takes the following form:
\begin{align*}
b^{t+1}(\phi_i)=\frac{p(\bar{\mu}_{H_i}^t|z_{H_{i}}^t,\phi_i)b^t(\phi_i)}
{\int p(\bar{\mu}_{H_i}^t|z_{H_{i}}^t,\phi_i)b^t(\phi_i)d\phi_i}.
\end{align*}
In practice, we find the above per-human update rule is still too costly to compute. Since the personality feature space is only 3D and the personality follows a specific distribution (determined by the mapping between RVO parameters and personality factors), we propose to discretize it into a set of $M$ representative personalities $\phi^{1\cdots M}$ via appropriate quantization.  Then, we conduct the discrete Bayesian update as:
\begin{align*}
b^{t+1}(\phi_i^j)=\frac{p(\bar{\mu}_{H_i}^t \mid z_{H_{i}}^t,\phi_i^j)b^t(\phi_i^j)}
{\sum_{j=1}^M p(\bar{\mu}_{H_i}^t \mid z_{H_{i}}^t,\phi_i^j)b^t(\phi_i^j)}.
\end{align*}

\subsection{Active Personality Probing}
At every timestep $t$, the robot only considers the observed part of the crowds and their corresponding personality beliefs, but for simplicity, our following description slightly abuse notations as if all humans were observed by the robot. Equipped with the learned navigation simulator and using the partial observations, the robot could predict the future trajectories of the joint system in a generative manner. With the belief over personality, however, the joint dynamic system becomes stochastic and estimating the exact distribution of trajectories is intractable. To alleviate the computational cost of prediction, we use the expected personality $\hat{\phi}_i=\sum_{j=1}^M \phi_i^j b(\phi_i^j)$ to predict the trajectories. 

Based on the trajectory prediction, we formulate robot planning as a non-linear optimization problem in continuous action space and solve it via Model Predictive Path Integral Control (MPPI). Our method is inspired by~\cite{information_gathering}, but extends it to the prediction problem of crowds with complex personality representation, which brings additional challenges. Following the concept of traditional Model Predictive Control (MPC), we design a cost function for the robot to compromise between various objectives. Our robot controller is expected to accomplish three goals simultaneously: exploring the environment, avoiding collisions, and most importantly, collecting information actively to help the robot better predict humans' personalities.

\paragraph{Exploring the Environment} We use a global planner to guide the robot to explore the environment so that it has more chances to encounter humans. The global planner provides the robot with a loopy path consisting of several goal positions and the robot would traverse to each goal position cyclically. We define the first term of robot cost function as follows:
\begin{align}
C_{\text{goal}}=\sum_{\Delta t=1}^{T_R}\exp\left(\alpha_{\text{goal}}\|p^{t+\Delta t}_{R}-g^{t+\Delta t}\|^2\right), \notag
\end{align}
where $T_R$ is the robot's control horizon, $\alpha_{\text{goal}}$ is the hyper-parameter, and $g^{t+\Delta t}$ is the next goal position at timestep $t+\Delta t$ given the robot's current location.

\paragraph{Collision Avoidance} We use another cost function to guide the robot to cautiously avoid collisions. To this end, we introduce the following cost:
\begin{align}
C_{\text{coll}}=\sum_{i=1}^N\sum_{\Delta t=1}^{T_R} \exp\left(-\alpha_{\text{coll}}\min(d_{\text{safe}},\|p^{t+\Delta t}_{R}-p^{t+\Delta t}_{H_i}\|^2)\right), \notag
\end{align}
where $d_{\text{safe}}$ is a user-defined safe distance threshold and $\alpha_{\text{coll}}$ is the hyper-parameter.

\paragraph{Active Information Gathering}
The main target of the robot is generating useful actions which could be beneficial for personality prediction. As mentioned before, the robot could infer the future system states when following a preplanned trajectory, so it could naturally obtain the tentative future belief $(b^{t+1}(\phi_i),...,b^{t+T_R}(\phi_i))$ for all observed humans via Bayesian update. Thus, we defined the information gain over humans' beliefs as the final term of robot cost function:
\begin{align}
C_\text{active}=\sum_{i=1}^N H\left(b^{t+T_R}(\phi_i)\right)-H\left(b^{t}(\phi_i)\right) \notag ,\quad H\left(b(\phi_i)\right)=-\textstyle\sum_{j=1}^Mb(\phi_i^j)\log\left(b(\phi_i^j)\right). \notag
\end{align}
The above cost term sums up the negative information gain of the belief over each observed human's personality. Optimizing this cost is equivalent to finding actions that could produce more valuable information, leading to a better personality estimation. Note that we will simultaneously optimize the information gain of all observed humans, which means we do not explicitly tell the robot which human needs more attention and the robot shall make a trade-off automatically when interacting with the crowds.

\paragraph{MPPI-based Controller} With the above cost terms, we formulate the robot control problem as a non-linear optimization:
\begin{align}
\mathop{\argmin}\limits_{\hat{\mu}^t_R}C=w_\text{goal}C_{\text{goal}}+w_\text{coll}C_{\text{coll}}+w_\text{active}C_{\text{active}},
\end{align}
where $w$ are different weight coefficients. We use an MPPI-based controller to find the optimal action sequence for our robot which could minimize the gross cost. In order to reduce the dimension of sampling space and improve the smoothness of trajectory, we sample $K$ Bezier curves $\mathcal{S}_{0...K}$ as the potential robot action trajectories. 
For each $\mathcal{S}_i$, we predict the corresponding human trajectories to estimate the cost $C(\mathcal{S}_i)$. We choose $\mathcal{S}^*$ with the lowest cost, and $\mathcal{S}^*$'s first timestep action $\hat{\mu}_{R}^t$ is then fed to the local collision avoidance network $\NeuralNet_f$ proposed in~\prettyref{sec:human_model_sec} to generate the final action $\mu_{R}^t = \NeuralNet_{f}(l_{R}^t, \hat{\mu}_{R}^t)$. It should be mentioned that our neural network based human behavior inference could use GPUs for parallelization and acceleration, which enables our optimization to perform nearly in real-time. \prettyref{alg:MPPI} outlines the MPPI-based optimization procedure.

\begin{algorithm}[ht]
\caption{MPPI-Based Robot Planner}
\label{alg:MPPI}
\begin{algorithmic}[1]
\Require{Number of samples $K$, control horizon $T_R$, current observed state $s^t$, current belief $b^t(\phi_{1\cdots N})$.}
\Ensure{Optimal action at current timestep $\mu_R^t$}
\State Sample $K$ Bezier curves $\mathcal{S}_{0\cdots K}$
\For{each robot control signal trajectory $\mathcal{S}_i$}
\State Sample on $\mathcal{S}_i$ to get robot control signals $\mu_R^{t\cdots t+T_R}$
\State Predict joint state trajectory $\{s^{t+1},\cdots,s^{t+T_R}\}$
\State Bayesian update to get tentative beliefs $b^{t+T_R}(\phi_{1\cdots N})$
\State Compute cost $C(\mathcal{S}_i)$
\EndFor
\State $\mathcal{S}^*\gets\argmin\limits_{\mathcal{S} \in {S}_{0\cdots K}}C(\mathcal{S})$
\State $\hat{\mu}_R^t\gets$first timestep of $\mathcal{S}^*$
\State Return $\NeuralNet_{f}(l_R^t, \hat{\mu}_R^t)$
\end{algorithmic}
\end{algorithm}

\revised{We test this planner and our personality predictor in simulation, and conduct user evaluations with typical adults and autistic participants separately, which could serve as a foundation for future active-robot-based ASD-specific interventions.}
\section{\label{sec:exp}Evaluations in Simulation}
In this section, we show the performance of each part of our approach in simulation settings. A continuous 2-D simulator is set up where the robot and humans are represented as circular holonomic agents with a physical radius of 5. We use ChAOS \cite{chaos} for 3D simulation, which is an open-source crowd animation software. We consider the scenario with only one robot and a crowd of different sizes. All our experiments are carried out on a PC with an 8-core, 2.30GHz Intel Core i7-11800H CPU and an Nvidia GeForce RTX 3060 GPU with 6GB memory. Our method is implemented using Python, and we use Pytorch for the training of neural networks and parallel trajectory prediction. The default parameters are set as: $T_H=5, T_R=10, d_H=100, d_R=200, \alpha_\text{goal}=0.003, \alpha_\text{coll}=0.1, d_\text{safe}=15, w_\text{goal}=0.0005, w_\text{coll}=0.04, w_\text{active}=1.0$. 

\begin{figure*}[ht]
\centering
\begin{minipage}[b]{0.3\linewidth}
    \includegraphics[width=0.90\linewidth]{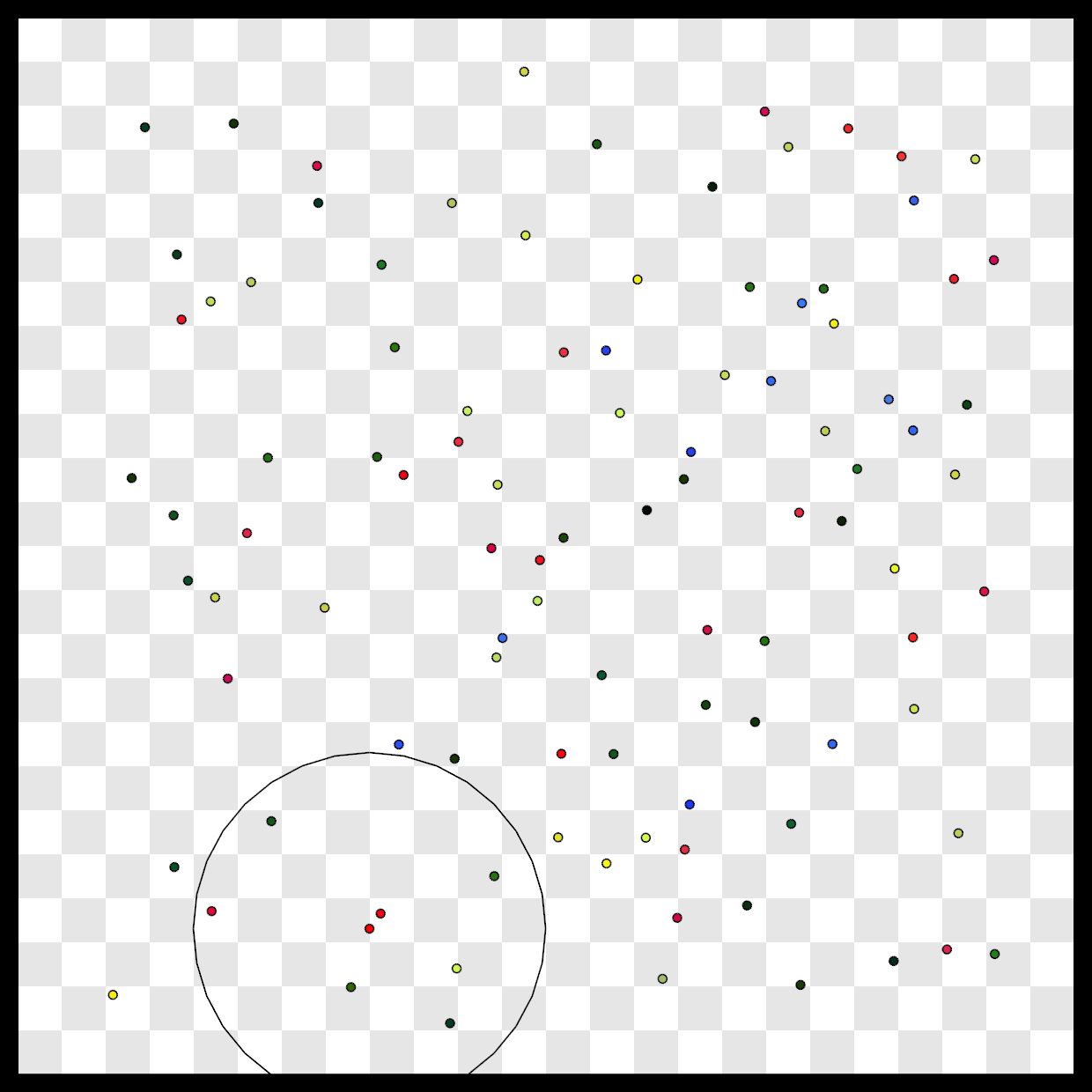}
\end{minipage}
\hspace{4px}
\begin{minipage}[b]{0.3\linewidth}
    \includegraphics[width=0.90\linewidth]{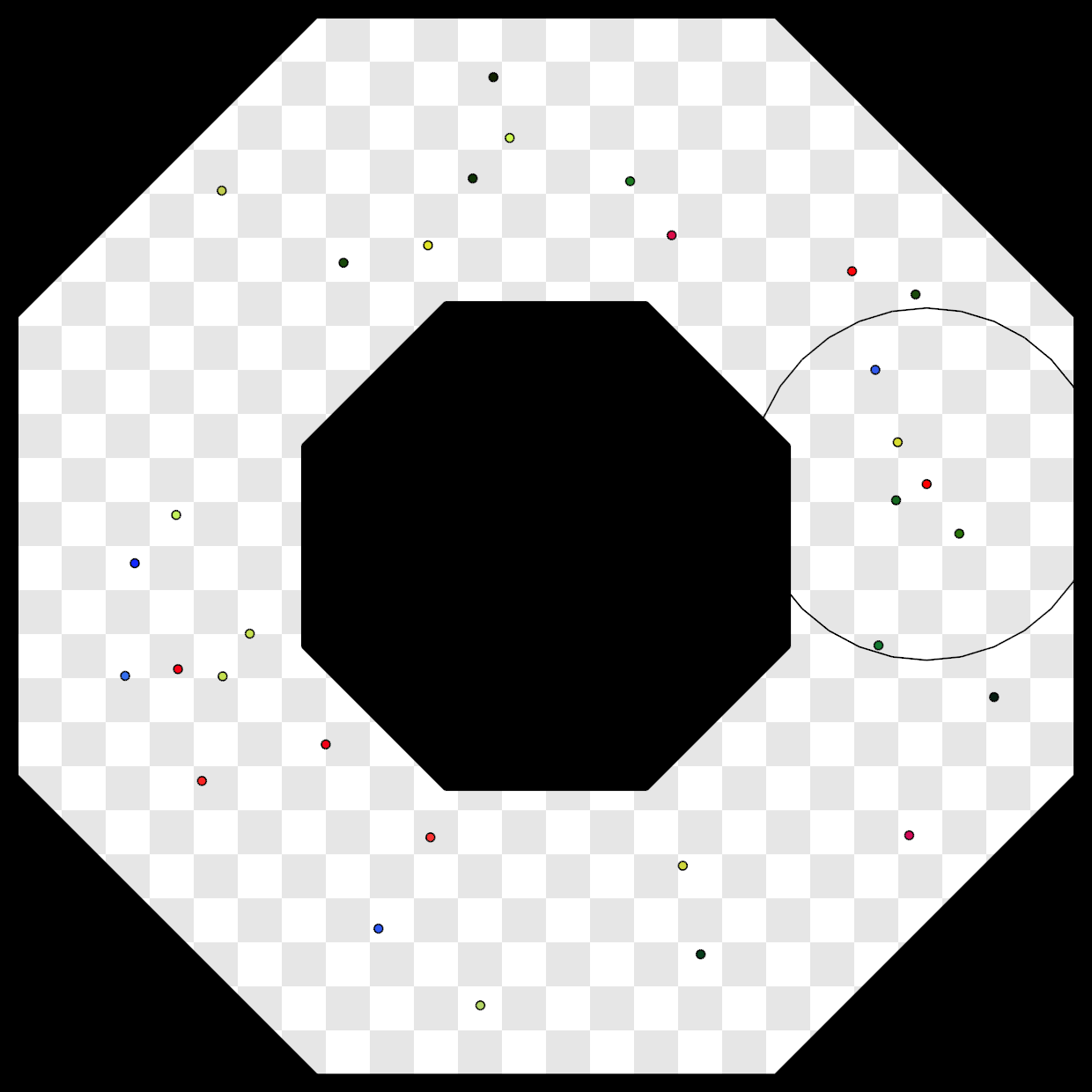}
\end{minipage}
\hspace{4px}
\begin{minipage}[b]{0.3\linewidth}
    \includegraphics[width=0.90\linewidth]{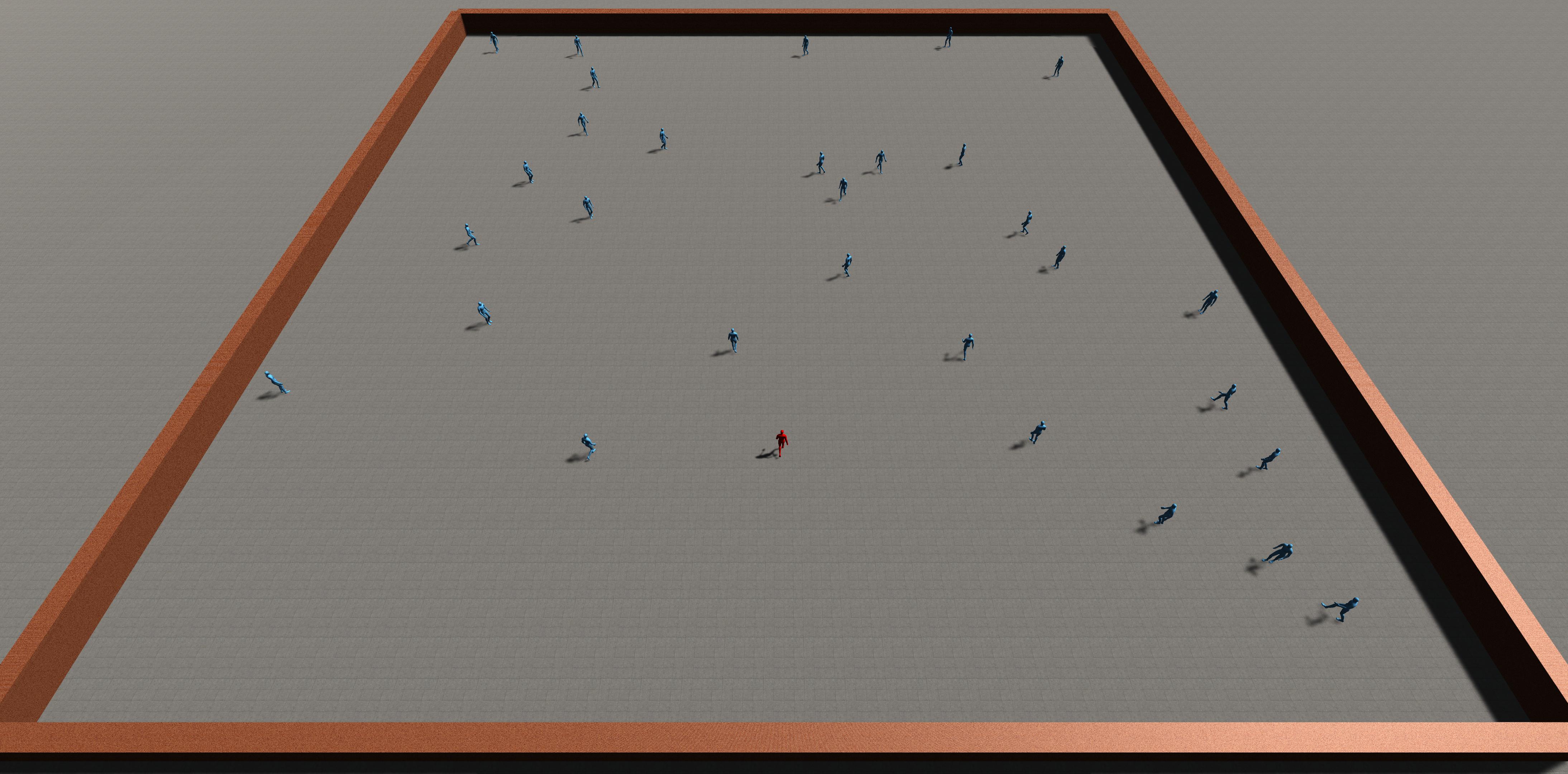}
    \includegraphics[width=0.90\linewidth]{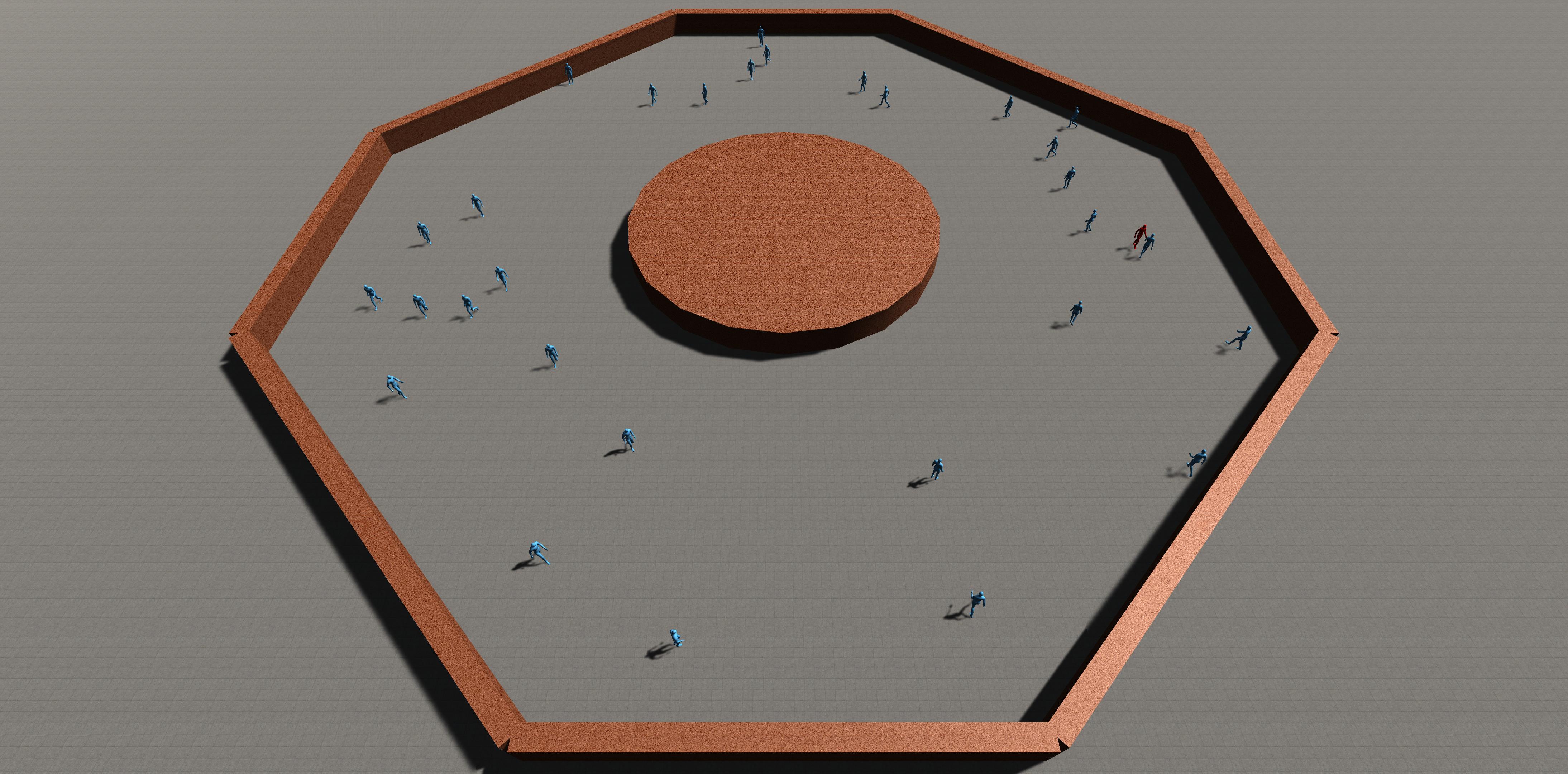}
\end{minipage}
\caption{Two environments used in experiments. Left: an empty room; Middle: a ring-like environment. Right: the corresponding environment in 3-D simulation. The red disk-shaped agent with circular sensor range in 2-D simulation and the red human in 3-D simulation represent the robot.}
\label{fig:scenarios}
\vspace{-5px}
\end{figure*}

~\prettyref{fig:scenarios} shows our 2-D and 3-D simulation scenarios. We use two closed environments which enable sustainable interactions: a large \textit{empty room} where humans move randomly and a \textit{ring-like} environment where humans keep moving around the circular space in the same direction. For the baseline, however, all other works for active internal state estimation are only considering a single human and one robot, to the best of our knowledge. These techniques cannot be trivially extended to multiple humans because of the computational cost and change of formulation. As a result, we mainly consider two algorithms with different cost settings, leading to completely different performances. The first is called an \textit{inactive controller}, where we remove the ``active'' term $C_{\text{active}}$ from our cost function and the robot fully relies on passive observations. We denote it as our baseline and we compare it with the full version of our approach: the \textit{active controller} that uses active probing actions. In addition, we carry out ablation studies to show the performance of each part of our method.

\subsection{Performance of Human Behavior Model}
\begin{figure}[ht]
\centering
\begin{subfigure}{0.42\linewidth}
    \centering
    \includegraphics[width=0.95\linewidth]{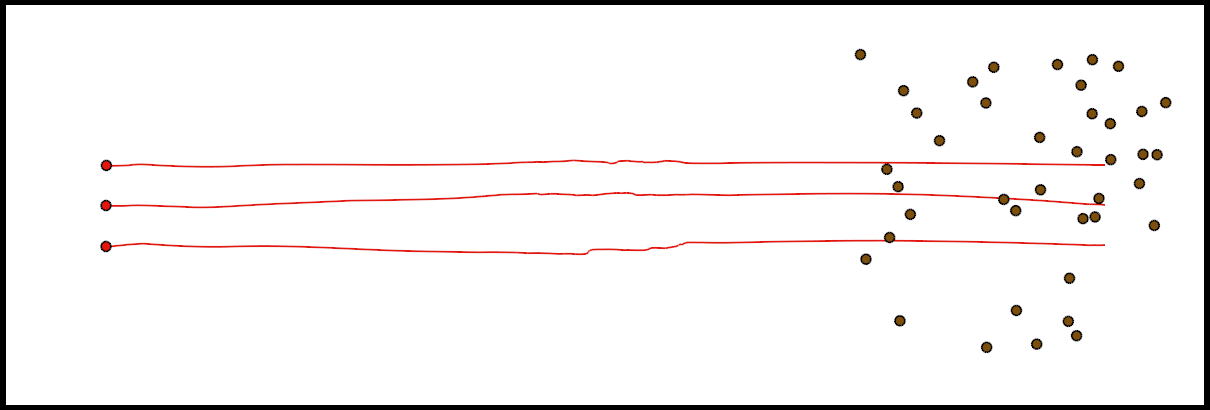}
    \caption{NN (Aggressive)}
\end{subfigure}
\hspace{6px}
\begin{subfigure}{0.42\linewidth}
    \centering
    \includegraphics[width=0.95\linewidth]{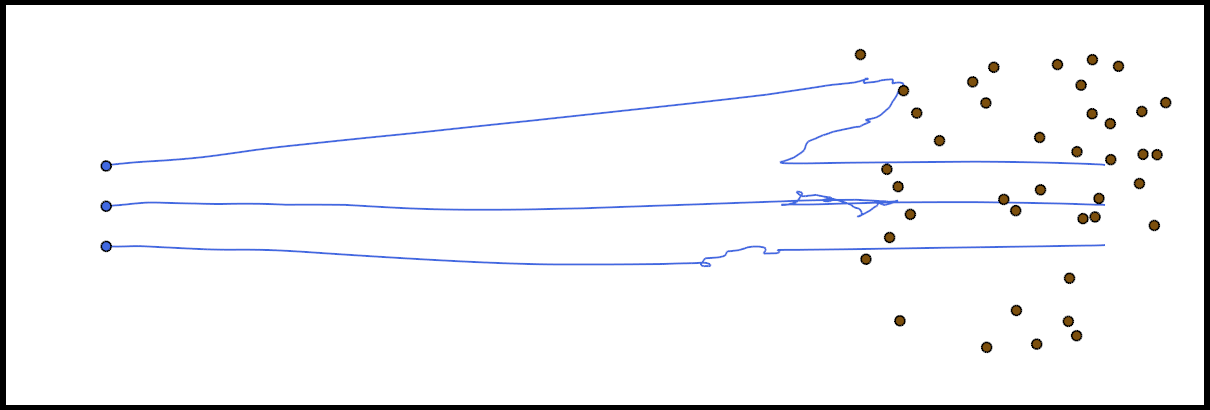}
    \caption{NN (Shy)}
\end{subfigure}
\begin{subfigure}{0.42\linewidth}
    \centering
    \includegraphics[width=0.95\linewidth]{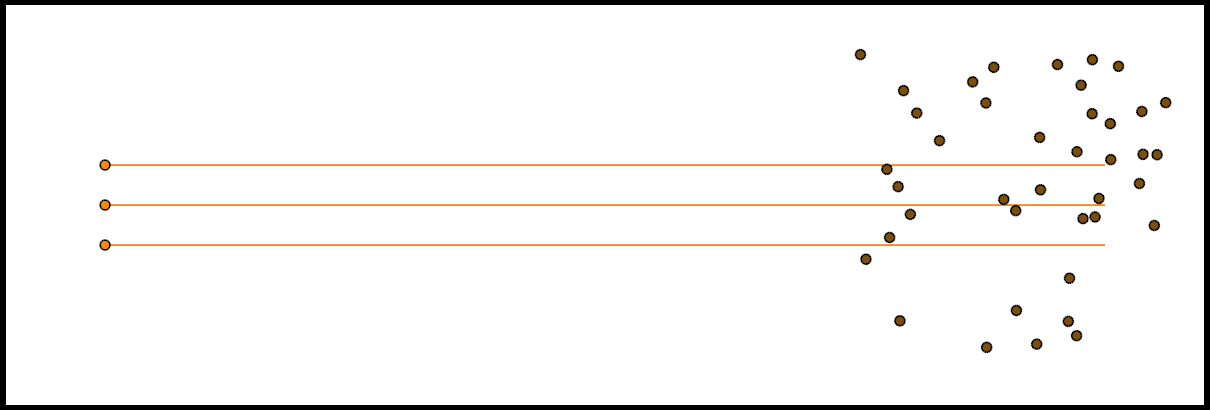}
    \caption{RVO (Aggressive)}
\end{subfigure}
\hspace{6px}
\begin{subfigure}{0.42\linewidth}
    \centering
    \includegraphics[width=0.95\linewidth]{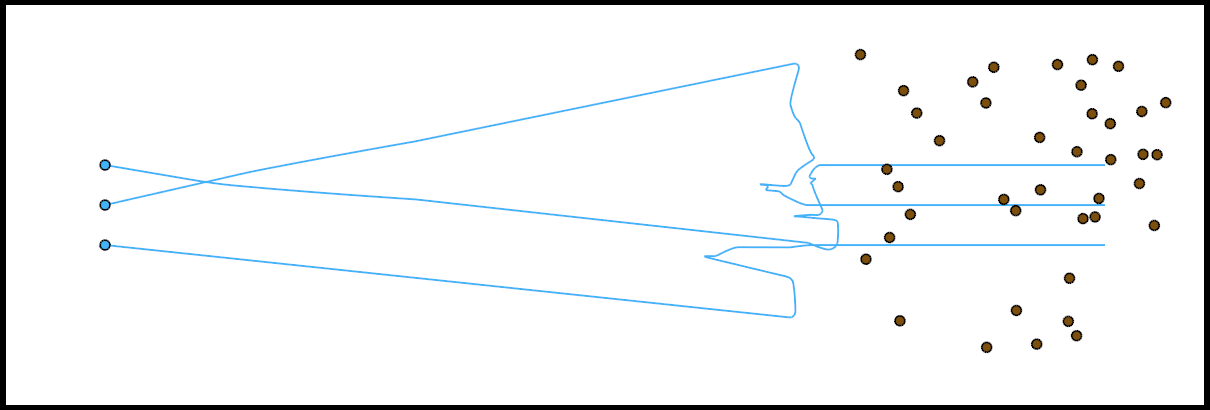}
    \caption{RVO (Shy)}
\end{subfigure}
\caption{Human behaviors generated by different models with different personalities. In each case, a crowd (brown) with the same control parameters is initially located on the left side and trying to move to the right side.}
\label{fig:human_behavior}
\vspace{-15px}
\end{figure}

Our learned human behavior could be understood as a soft, differentiable version of RVO. \prettyref{fig:human_behavior} illustrates 3 human behaviors generated by RVO and our learning-based method, when facing the same crowd situation (the crowd is moving from left to right). We experiment with two types of personalities: ``aggressive'' and ``shy''. As shown in~\prettyref{fig:human_behavior}, aggressive humans controlled by both NN and RVO show strong aggressiveness, which have less avoidance behaviors when facing the crowd. Conversely, shy people tend to avoid contact with others and try to dodge or wait. The humans controlled by NN exhibit a similar tendency, which implies our neural-network-based human behavior model could provide explainable personality-conditioned behavior.

\subsection{\label{sec:psn_est}Passive versus Active Robot Behaviors}
In this part, we will demonstrate the remarkable performance of our active controller by comparing it with the inactive baseline. We will demonstrate two main aspects:
\begin{itemize}
\item Our method results in reasonable interaction behaviors;
\item Our method results in more interactions which leads to more accurate personality estimation.
\end{itemize}

\begin{figure*}[ht]
\centering
\begin{subfigure}{0.47\linewidth}
\centering
    \fbox{\includegraphics[width=0.45\linewidth, trim=100 30 0 210,clip]{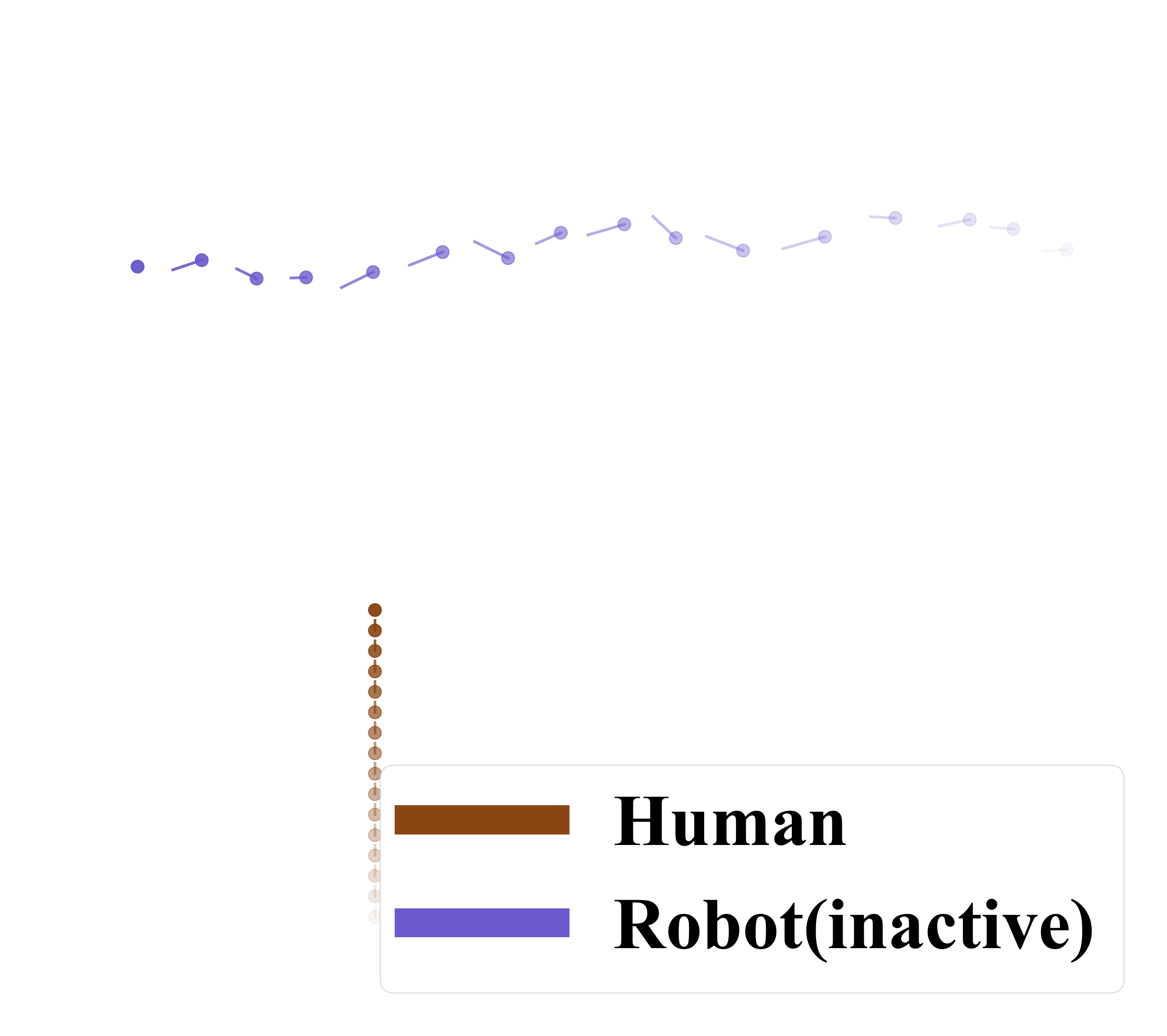}}
    \fbox{\includegraphics[width=0.45\linewidth, trim=100 30 0 210,clip]{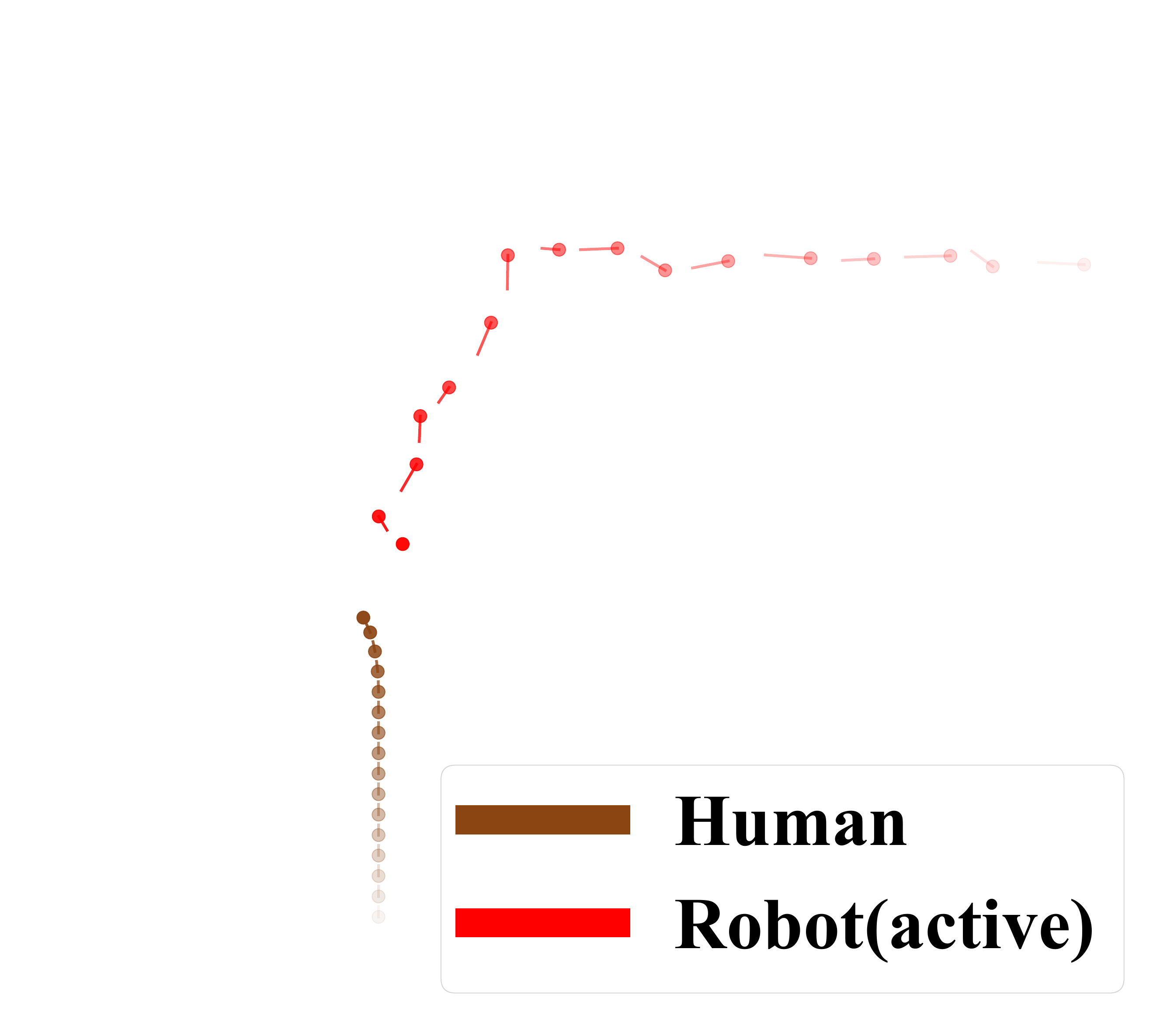}}
    \includegraphics[width=0.482\linewidth, trim=200 0 100 40,clip]{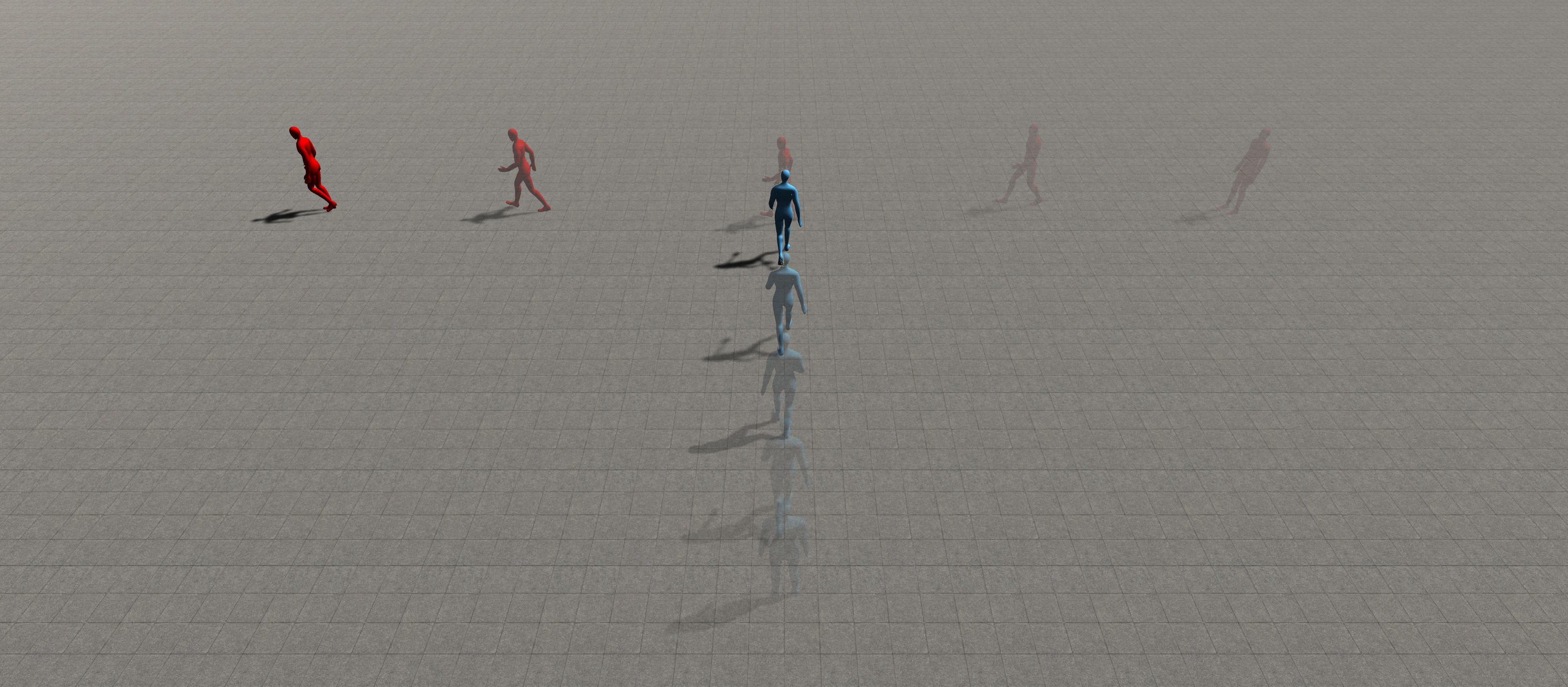}
    \includegraphics[width=0.482\linewidth, trim=200 0 100 40,clip]{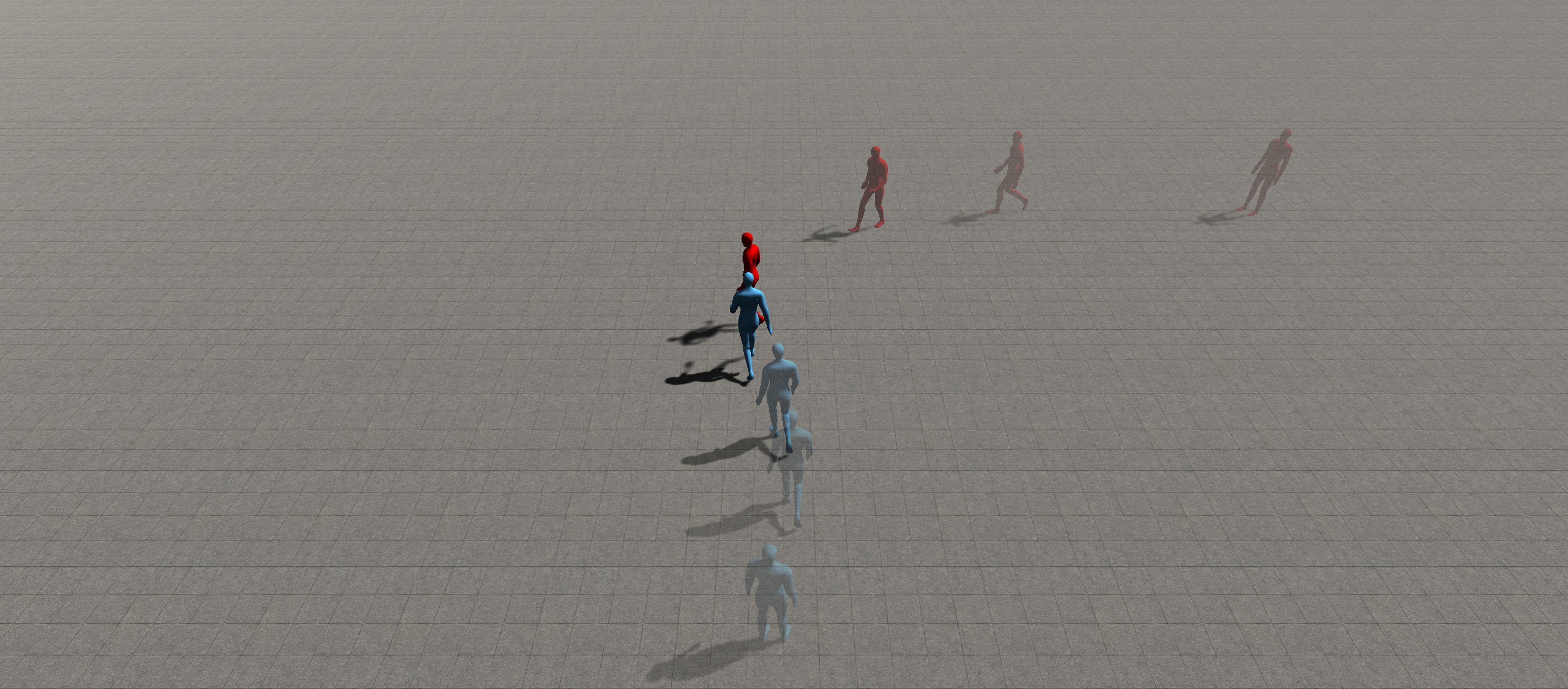}
    \caption{Robot behaviors generated by the inactive controller (left) and the active controller (right), respectively.}
    \label{fig:interactive_behavior}
\end{subfigure}
\hspace{10px}
\begin{subfigure}{0.46\linewidth}
    \centering
    \hspace{0px}\includegraphics[width=0.55\linewidth, trim=70 110 70 100,clip]{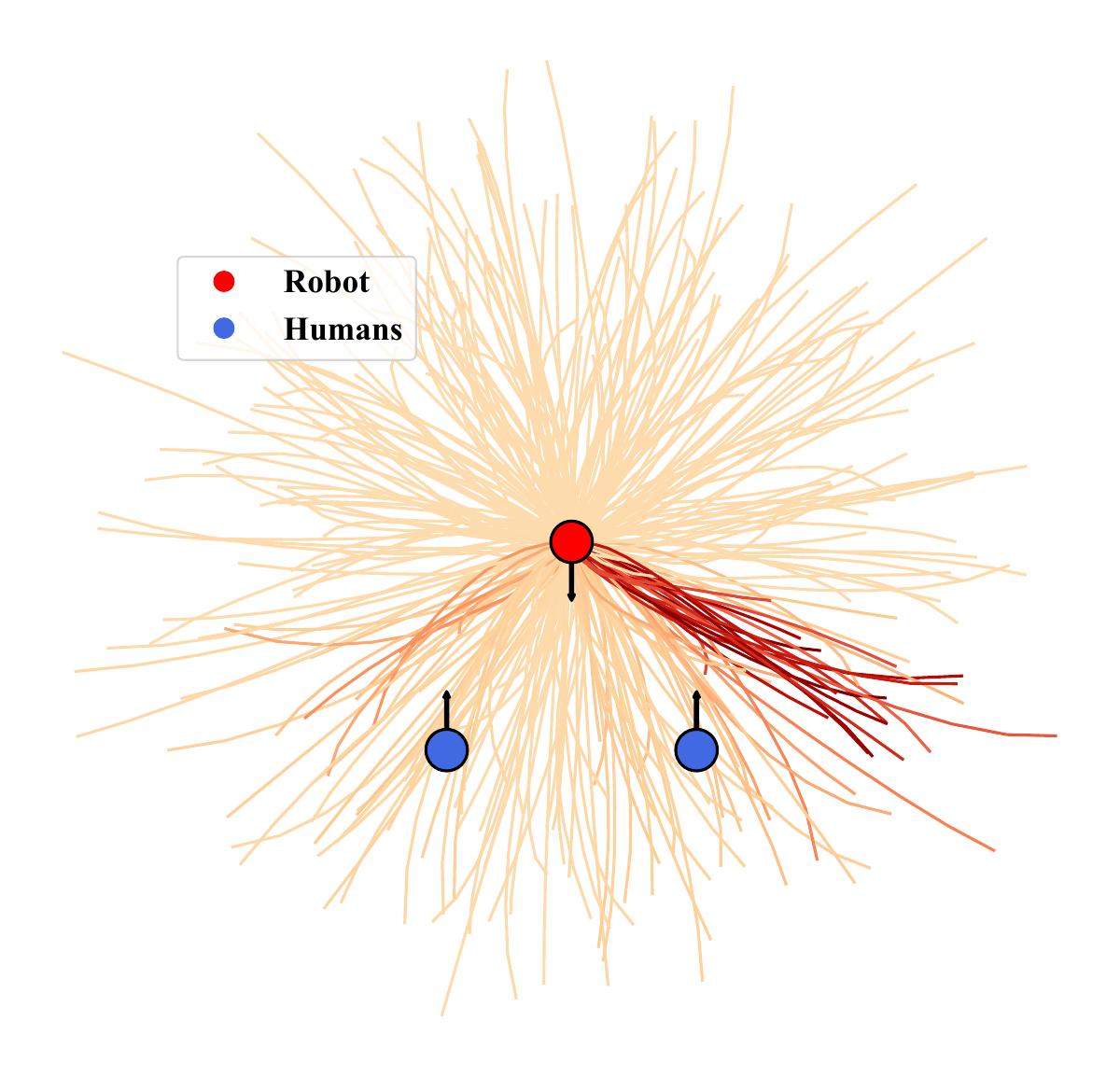}
    \hspace{0px}\includegraphics[width=0.335\linewidth,trim=770 140 770 60,clip]{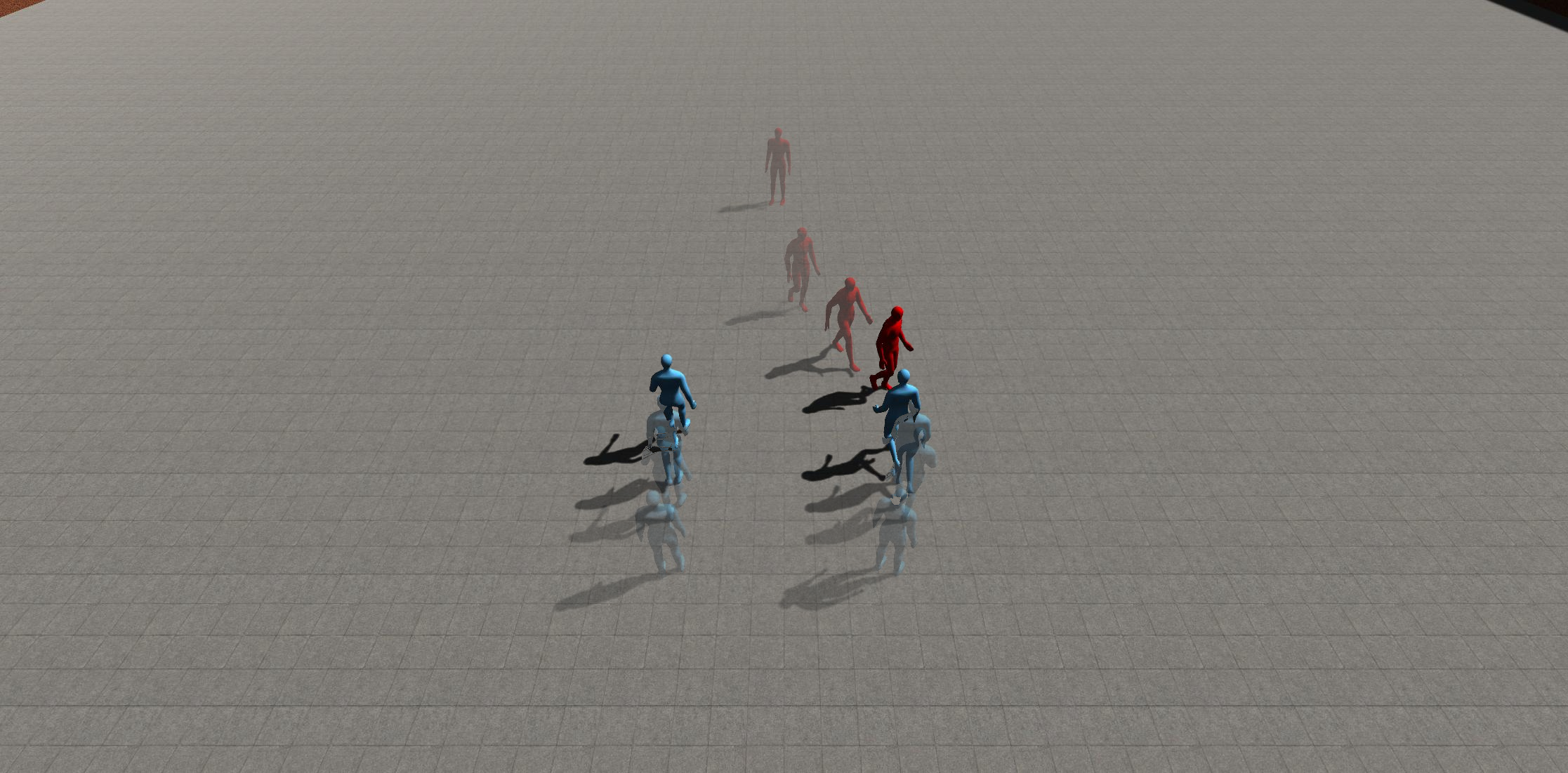}
    \caption{\textbf{Left}: The sampled trajectories of the robot (red) when facing two humans (blue), one is familiar to it (bottom left) and another is a stranger (bottom right). Warmer color corresponds to lower $C_\text{active}$. The robot tends to turn right and interact with the stranger. \textbf{Right}: The same case in 3-D simulation.} 
    \label{fig:exploration}
\end{subfigure}
\caption{Interactive behavior of the robot.}
\vspace{-15px}
\end{figure*}

\paragraph{Interactive Behavior}
\prettyref{fig:interactive_behavior} plots the trajectories generated by the inactive controller (left) and our active controller (right), respectively, when one human and the robot try to cross an intersection in different directions. Compared with the inactive robot moving straightforward, the active robot changes its moving direction in advance and tries to interact with the human actively, which triggers more human reaction to serve as the basis for personality prediction.

We further find that the active controller can automatically guide the robot to focus on humans ``stranger'' to the robot, i.e., of whom the entropy of belief is high. If the belief on one human's personality is quite low, the robot will find that the benefits from interacting with that human is also low, because interacting with him or her will bring little entropy changes. We illustrate this process in~\prettyref{fig:exploration}. The robot (red) is moving downward and observes two humans (blue) moving in the opposite direction. The human on the left is someone it is ``familiar'' with, i.e., the entropy of belief is low, and the right human is someone the robot knows nothing about. Our active controller will then sample several trajectories and compute the information gain brought by each trajectory. \prettyref{fig:exploration} illustrates the sampled trajectories and the corresponding information gain, i.e., $C_{\text{active}}$, in which warmer color represents lower cost. It is clear that the robot is more inclined to interact with the unfamiliar human on the right, which could be seen as an implicit greedy exploration strategy.

\begin{figure*}
\centering
\includegraphics[width=0.40\linewidth]{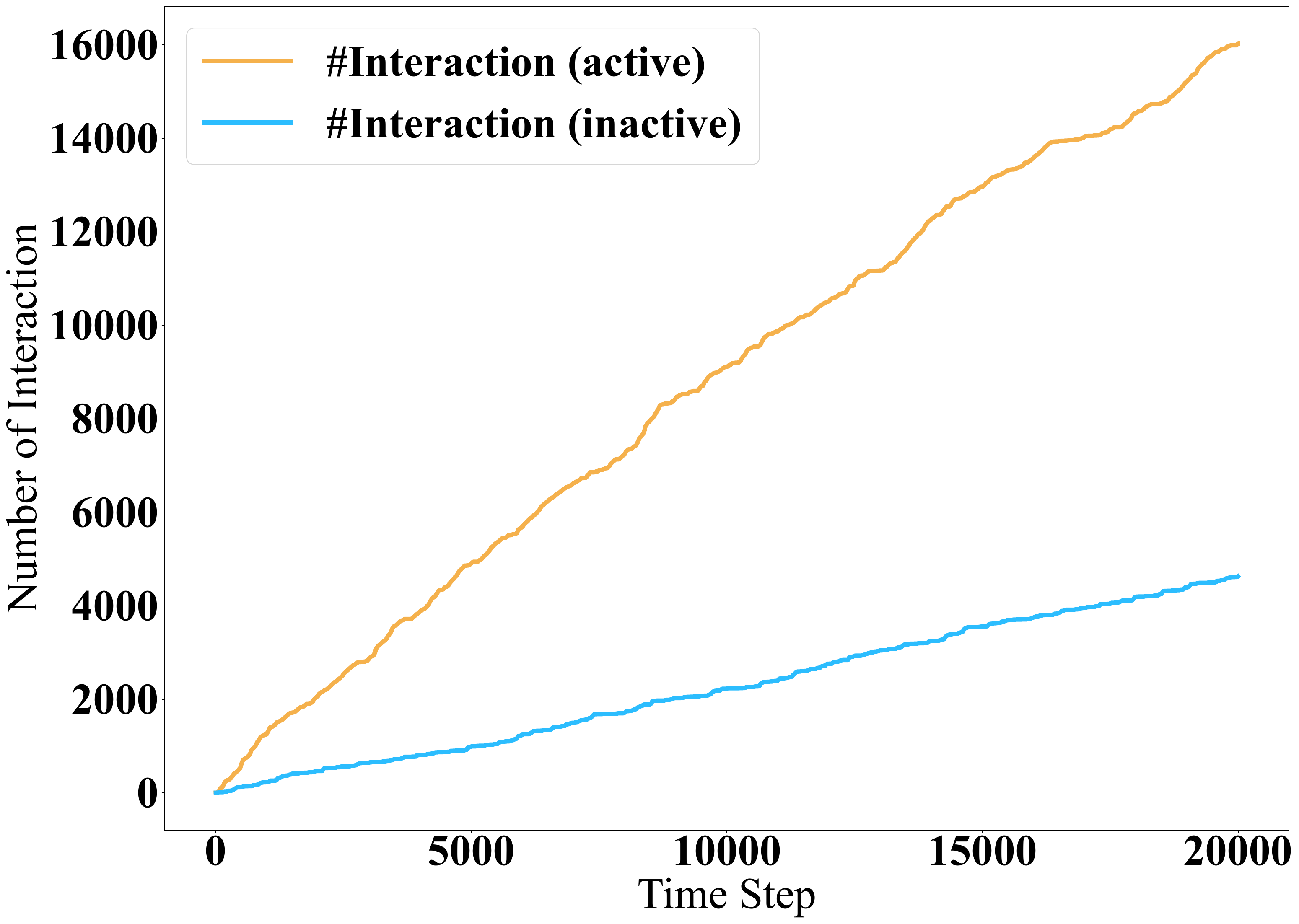}
\caption{The number of observed interactions changing history under active and inactive controllers.}
\label{fig:interactive_cnt}
\vspace{-10px}
\end{figure*}

\paragraph{Number of Interactions}
We visualize the number of interactions that the robot observed within $20,000$ time steps under the active controller and inactive controller, respectively. The experiment is carried out in the room environment with 30 humans. \prettyref{fig:interactive_cnt} shows how the number of interactions changes over time for both algorithms. It could be clearly seen that the active controller brings more robot-human interactions, which we believe is more likely to contain useful information for better human personality estimation. It is worth noting that, while our active controller leads to much more interactions, it still well guarantees collision avoidance. Specifically, we count the collision times when executing using both active and inactive controllers, where we judge a collision happens if the distance between the robot and a human is less or equal to twice the agent's physical radius. We have observed 0 collisions in both two methods, which verifies the robust collision handling of our system. \revised{In practice, we can balance the navigation purpose and information gathering by adjusting the weights of the optimization objectives. For instance, increasing the weights of ``collision avoidance'' and ``moving to goal'' can prioritize general navigation over active interaction behaviors. This allows the robot to simulate the effect of ``passing by'' humans, ensuring \textit{human comfort} is considered during interactions.}

\paragraph{Performance of Personality Estimation}
We use two metrics to quantitatively evaluate these two methods. We define \textit{prediction error (PE)} as the \textit{Mean Squared Error (MSE)} between the expected personality $\hat{\phi}_i$ and ground truth personality ${\phi^\star_i}$. We further define the \textit{belief entropy (BE)} as the entropy of the updated belief, which could be represented as $H\left(b(\phi_i)\right)$. The number of interactions that the robot observes is also counted where we assume that an interaction happens when the distance between two agents is smaller than $0.3d_H$. We run experiments with $1$ robot and $1,5,10,30,100$ humans for both environments, and the groundtruth personalities of humans are sampled randomly. 

All experimental results are summarized in~\prettyref{tab:all_prediction_results_circle} and~\prettyref{tab:all_prediction_results_empty}. It is evident that the active controller brings more robot-human interactions (``Itr'' is an abbreviation for ``Interaction''). \revised{We also observe that the number of interactions initiated by the inactive controller is slightly higher than those by the active controller in the ring-like scenario with 100 humans, likely due to random variations. This suggests that the environment with 100 humans is quite crowded, resulting in numerous natural interactions.} Meanwhile, the active controller always performs better than the inactive one in both PE and BE. The performance of active controller does not show significant correlation with crowd density. By comparison, the PE and BE obtained by the inactive controller reduces with the increasing number of humans, which is marginally higher than the active method when the number of humans is 100. This implies that our active controller could be more effective in sparse environments, i.e., large scenario with few humans. The reason behind this is also easy to understand: Sparse environment can induce less passive interaction between humans, which requires more active behaviors of the robot for information gathering. It should also be mentioned that, in practice, we inject Gaussian noise $\mathcal{N}(0, \sigma^2=(\pi/6)^2)$ to the output of $\NeuralNet_F$ to mimic the system error of personality-driven navigation model, which is not injected in the trajectory estimation procedure in~\prettyref{alg:MPPI}. The above results indicate that our method naturally adapts to this noise. 

\begin{figure*}[ht]
\centering
\begin{subfigure}{0.48\linewidth}
    \centering
    \includegraphics[width=0.49\linewidth]{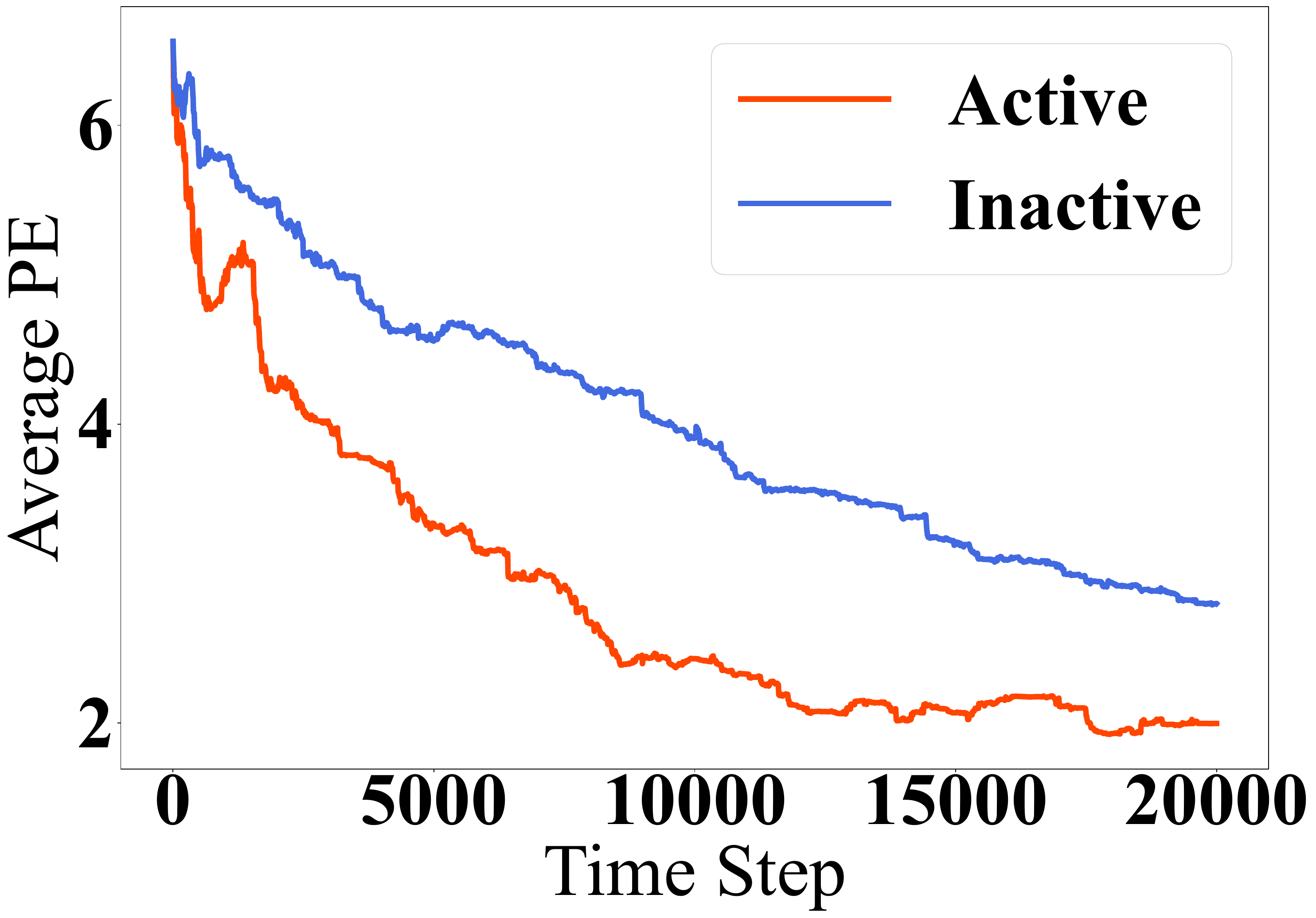}
    \includegraphics[width=0.49\linewidth]{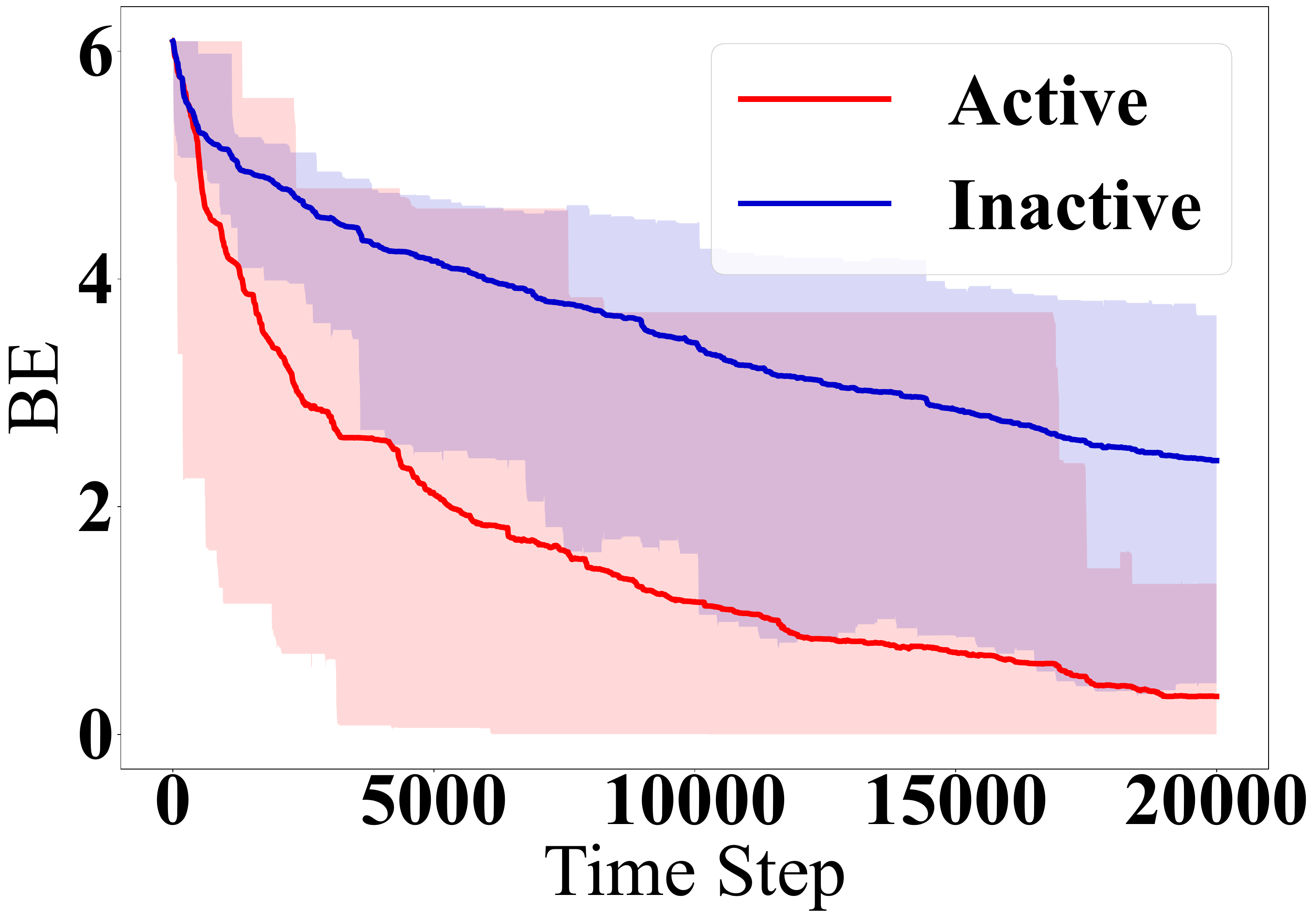}
    \caption{The average PE (left) and BE (right) obtained by active and inactive controllers, respectively. The shaded part represents the BE range of all human individuals.}
    \label{fig:prediction_comparison}
\end{subfigure}
\hspace{5px}
\begin{subfigure}{0.48\linewidth}
    \centering
    \includegraphics[width=0.49\linewidth]{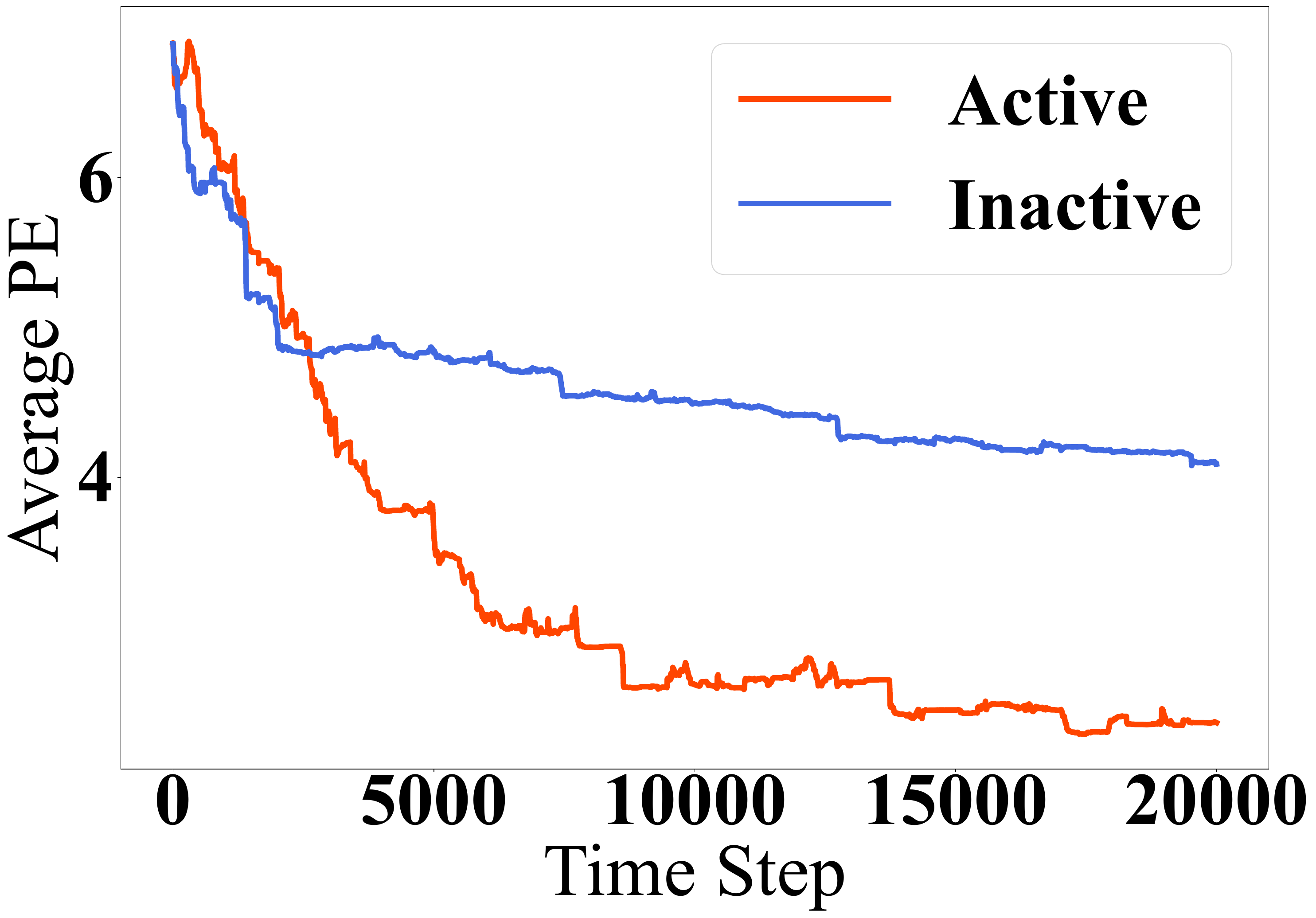}
    \includegraphics[width=0.49\linewidth]{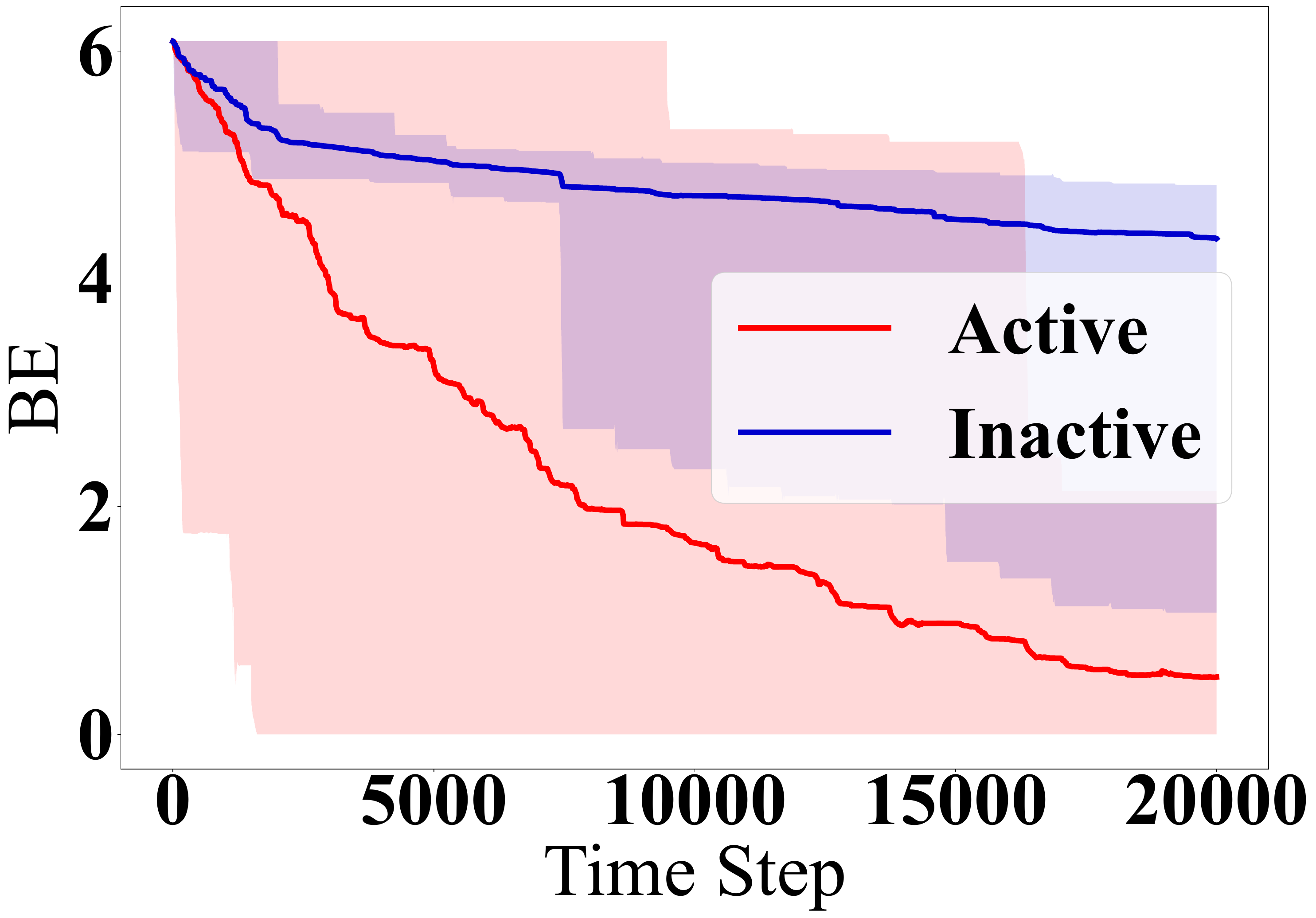}
    \caption{Same as (a), but enlarge the scenario by 3 times. The active method converges well, but the performance of the inactive method becomes worse.}
    \label{fig:prediction_comparison_3L}
\end{subfigure}
\caption{The changes of average PE and BE in ring-like cases.}
\vspace{-10px}
\end{figure*}

To clearly visualize the above experimental results, we take the ring-like scenario with 30 humans as an example and plot the change of PE and BE. As shown in~\prettyref{fig:prediction_comparison}, our active controller exhibits faster convergence and lower final value than the inactive controller in terms of PE. In addition, the BE of the active method also drops faster, which means the additional interactions do provide more useful information for reducing prediction uncertainty. 

As mentioned before, our active controller could be more effective in sparse environments. We perform an additional experiment in the same scenario as \prettyref{fig:prediction_comparison} but enlarge the environment by three times, of which results are illustrated in~\prettyref{fig:prediction_comparison_3L}. Our active controller leads to consistently fast decrease of PE and BE, but the inactive method has a worse performance due to inefficient interaction. 

\begin{minipage}{\linewidth}
\hspace{-6px}
\begin{minipage}[t]{0.49\linewidth}
    \begin{table}[H]
    \centering
    \setlength{\tabcolsep}{4px}
    \begin{tabular}{cccccc}
    \toprule
    \#Human & 1 & 5 & 10 & 30 & 100 \\
    \midrule
    PE active & 1.674 & 2.195 & 1.882 & 1.997 & 1.847 \\
    BE active & 0.051 & 0.003 & 0.064 & 0.334 & 0.588 \\
    \#Itr active & 2029 & 3655 & 11804 & 19024 & 96275 \\
    PE inactive & 3.456 & 2.834 & 2.724 & 2.799 & 2.130 \\
    BE inactive & 4.077 & 2.931 & 3.189 & 2.405 & 1.191 \\
    \#Itr inactive & 82 & 794 & 1690 & 12137 & 103012 \\
    \bottomrule
    \end{tabular}
    \caption{\label{tab:all_prediction_results_circle}{Results in the ring-like scenario.}}
    \end{table}
\end{minipage}
\begin{minipage}[t]{0.49\linewidth}
    \begin{table}[H]
    \centering
    \setlength{\tabcolsep}{4px}
    \begin{tabular}{cccccc}
    \toprule
    \#Human & 1 & 5 & 10 & 30 & 100 \\
    \midrule
    PE active & 1.241 & 2.301 & 1.985 & 1.965 & 2.097 \\
    BE active & 0.418 & 0.560 & 0.838 & 0.625 & 0.583 \\
    \#Itr active & 791 & 2802 & 5909 & 16026 & 70643 \\
    PE inactive & 2.591 & 3.006 & 2.993 & 2.825 & 2.224 \\
    BE inactive & 4.190 & 3.392 & 3.647 & 2.337 & 0.998 \\
    \#Itr inactive & 26 & 273 & 714 & 4628 & 53590 \\
    \bottomrule
    \end{tabular}
    \caption{\label{tab:all_prediction_results_empty}{Results in the square scenario.}}
    \end{table}
\end{minipage}
\end{minipage}

\subsection{Running Time}
We have calculated the running time of our algorithm when generating trajectory for the robot as well as optimizing personality estimation for multiple humans. \revised{Compared with the prior work~\cite{information_gathering} that requires approximately 0.3s to solve a 5-horizon-length optimization step for 1 human, our algorithm uses a longer horizon length of $T_R=10$ and solves each time step of the optimization in 0.075s, 0.192s, and 0.463s for 1, 10, and 30 humans respectively, which is real-time performance.} Due to the neural network's parallelism advantage, the running time of our method does not increase linearly with the number of humans.

\subsection{Visualizations of Personality Estimation}
\begin{figure*}[ht]
\centering
\begin{subfigure}{0.56\linewidth}
    \centering
    \includegraphics[width=0.48\linewidth]{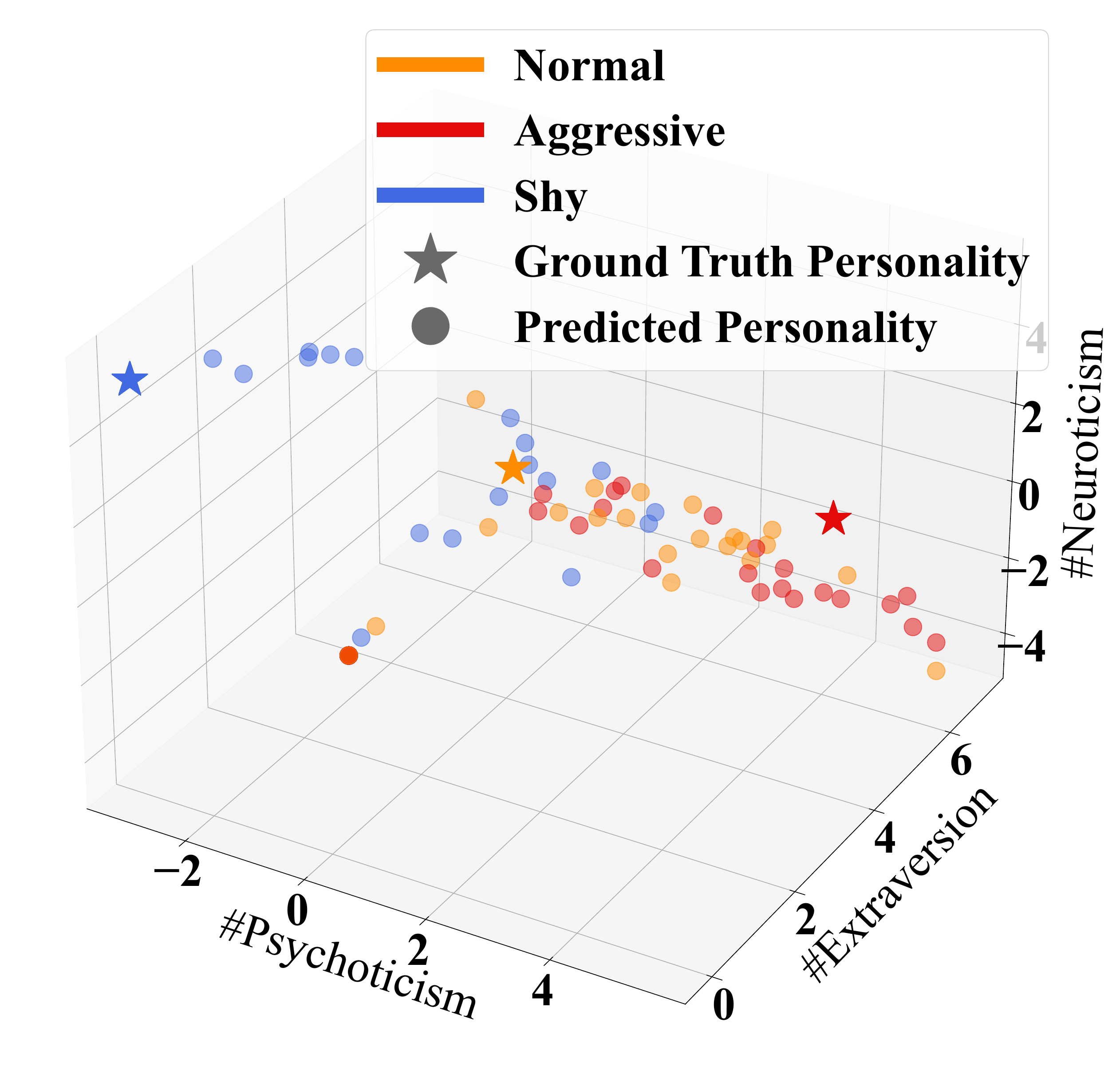}
    \includegraphics[width=0.48\linewidth]{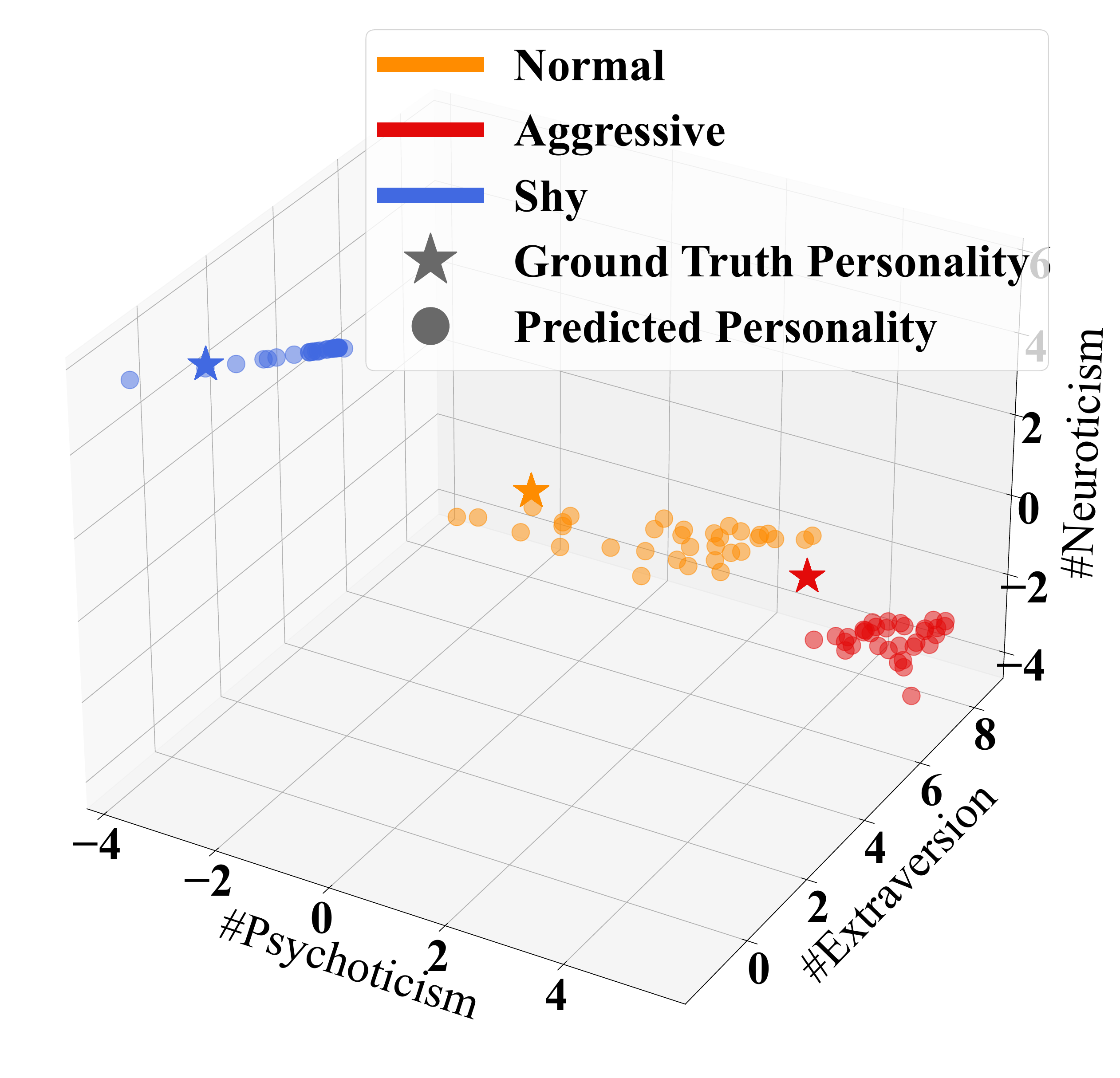}
    \caption{The prediction effect at the early stage of our algorithm (left) and after many interactions (right). The predicted personalities of 3 kinds of humans are plotted using scatters and the ground truth is represented as stars. In the beginning, the robot can not distinguish humans with different personalities; but in the later stage, the prediction shows a clear boundary.}
    \label{fig:psn_prediction_scatter}
\end{subfigure}
\hspace{5px}
\begin{subfigure}{0.40\linewidth}
    \centering
    \includegraphics[width=0.98\linewidth]{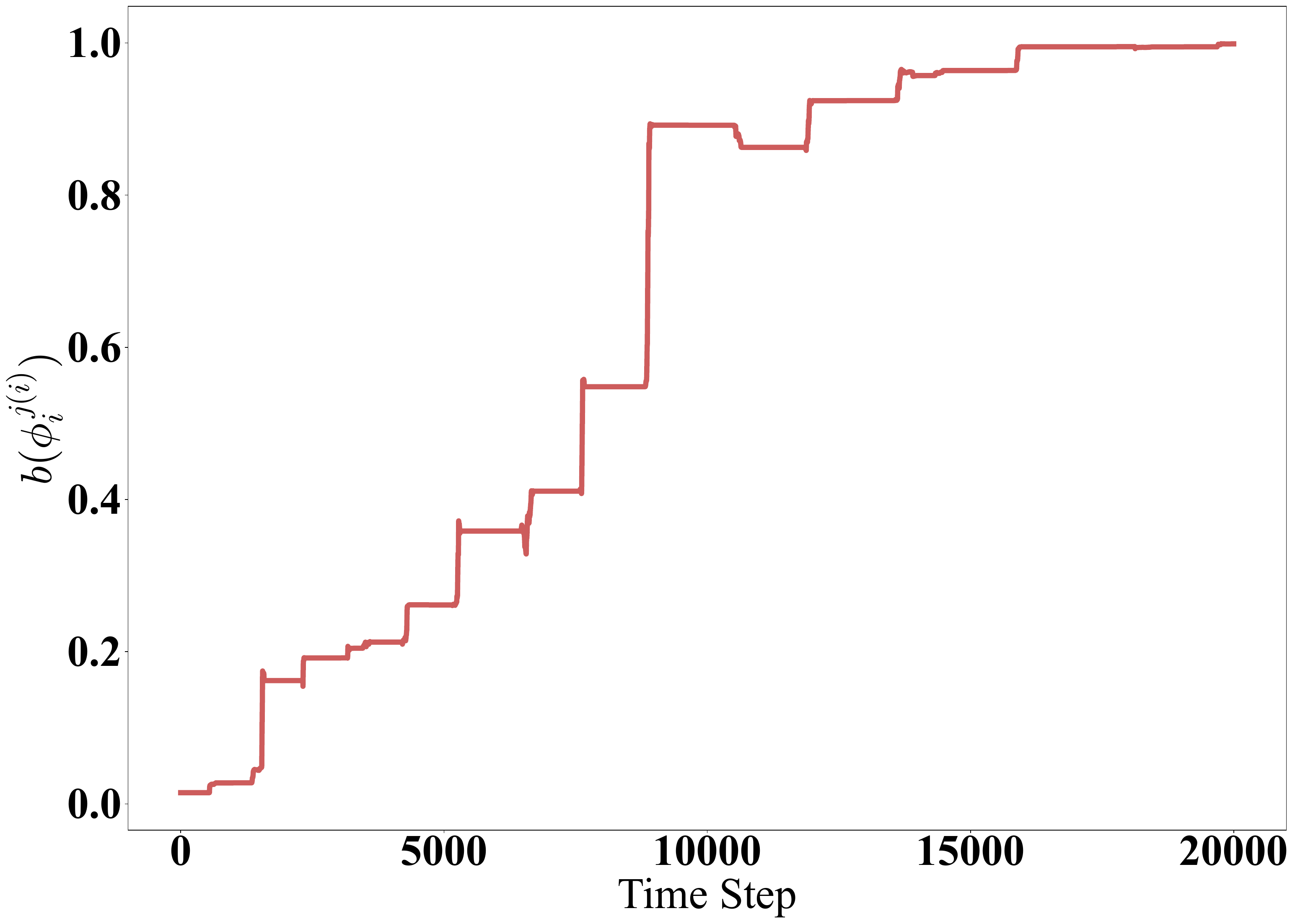}
    \caption{Changing process of the belief of human $H_i$'s personality. The probability of correct personality prediction continues to rise and becomes close to $1$ after $15000$ time steps.}
    \label{fig:belief_single_H}
\end{subfigure}
\caption{Visualizations of personality prediction process.}
\end{figure*}

To further analyze the influence of our active controller on each human, we select one human and illustrate how the belief of his/her personality changes during the experiment. Let human $H_i$'s ground truth personality be $\phi_i^\star$ and the closest discretized personality be $\phi_i^{j(i)}$, we plot the changing process of $b\left(\phi_i^{j(i)}\right)$ in~\prettyref{fig:belief_single_H}. It could be seen that the belief keeps increasing, which means the robot's prediction of $\phi_i^\star$ is becoming more and more accurate.

To better visualize the overall prediction result, we perform an additional experiment that uses the empty scenario with $100$ humans, but the personalities of the crowds are set to only three categories: 33 aggressive, 33 normal, and 34 shy. We use our active method for robot control. \prettyref{fig:psn_prediction_scatter} illustrates the estimation results at different stages. At the beginning (left), the expected personalities of most humans is inaccurate. Over a sufficiently long time (right), the prediction gets very close to the groundtruth.

\subsection{Privacy Security}
\begin{figure*}
\centering
\includegraphics[width=0.40\linewidth]{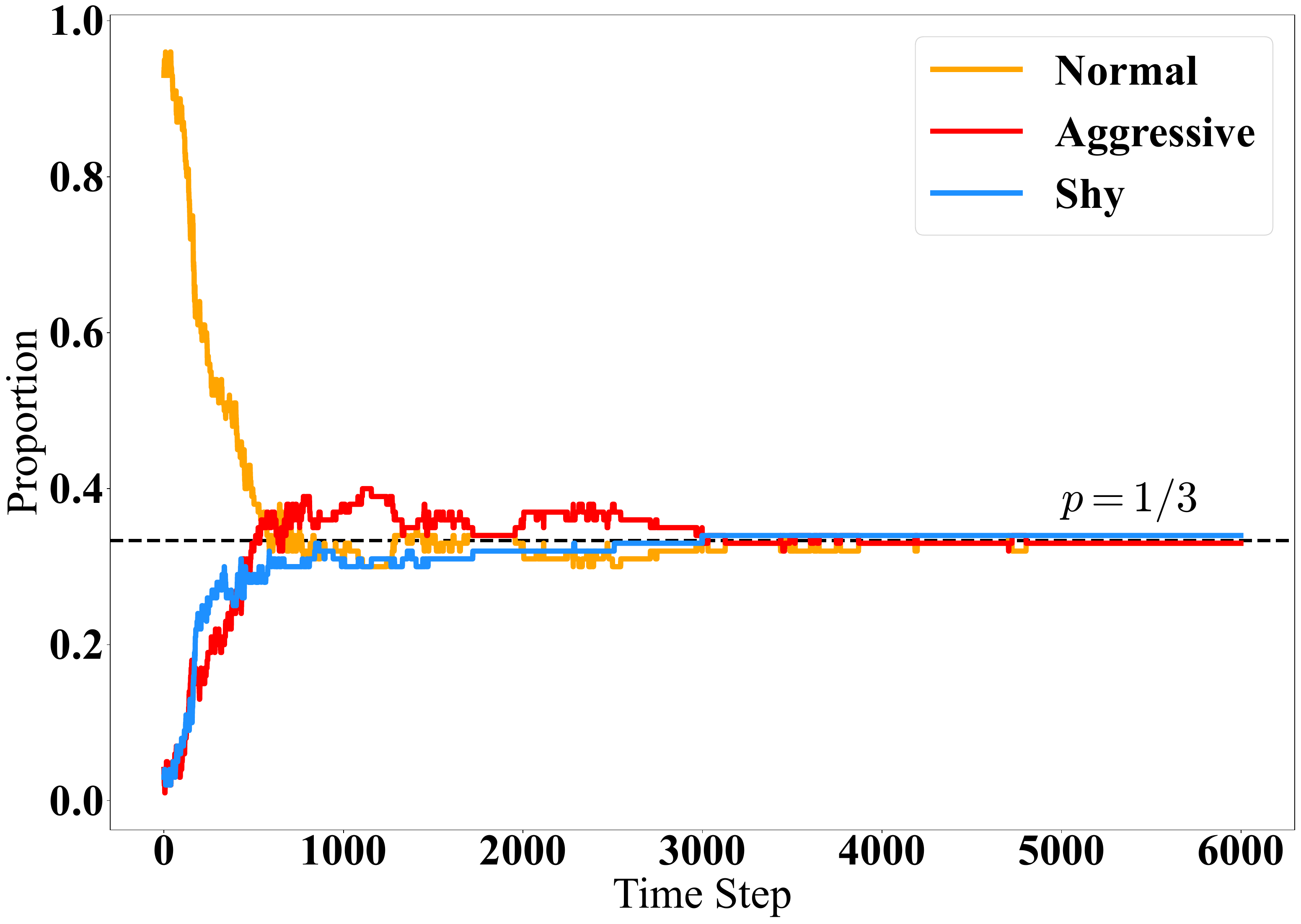}
\caption{Proportion changing history of each personality. The dotted line represents the real situation, i.e. each kind of personality accounts for one third.} 
\label{fig:whole_crow_psn}
\vspace{-10px}
\end{figure*}
Although we update beliefs for each human in the prediction process, we can choose to only retain the personality distribution over the entire crowd, securing the sensitive human's personality information. Following the setting that involves 100 humans with 3 kinds of personalities, for example, we assume that the robot has a priori knowledge of and carry out inference only on the three personality categories, and classify each human's personality. Then, we compile statistics on the proportion of each category in crowds at different stages and discard other information. As shown in~\prettyref{fig:whole_crow_psn}, the robot knows little about the personality distribution of the crowd initially. With the increase of exploration, the estimation of the proportions gradually becomes accurate, i.e., about one third of each category. As a result, we can still obtain valuable crowd personality distribution without disclosing human identity.

\subsection{\label{sec:model_corr}Robustness to Model Corruption}
\begin{figure*}[ht]
\centering
\begin{subfigure}{0.67\linewidth}
    \centering
    \includegraphics[width=0.48\linewidth]{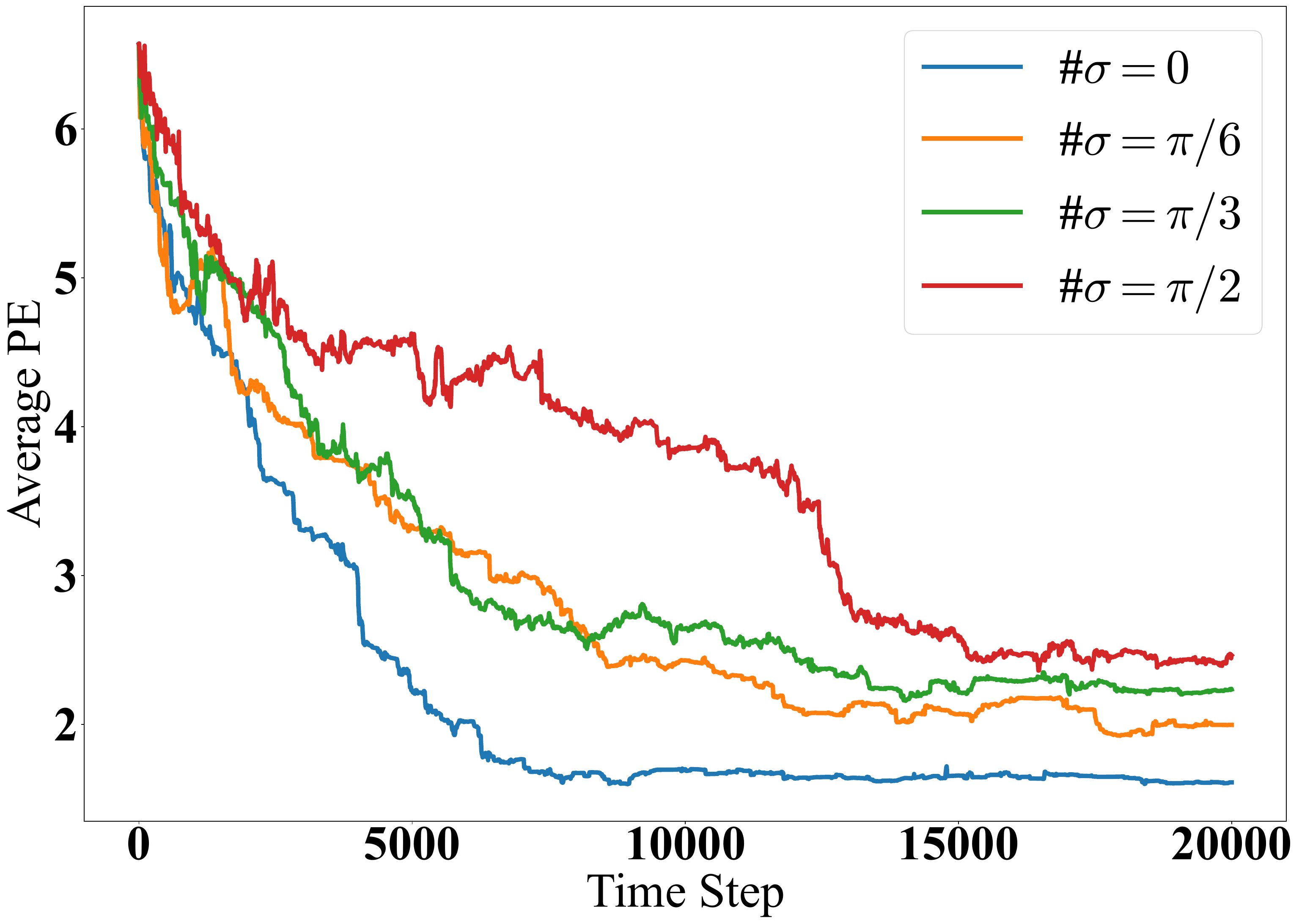}
    \includegraphics[width=0.48\linewidth]{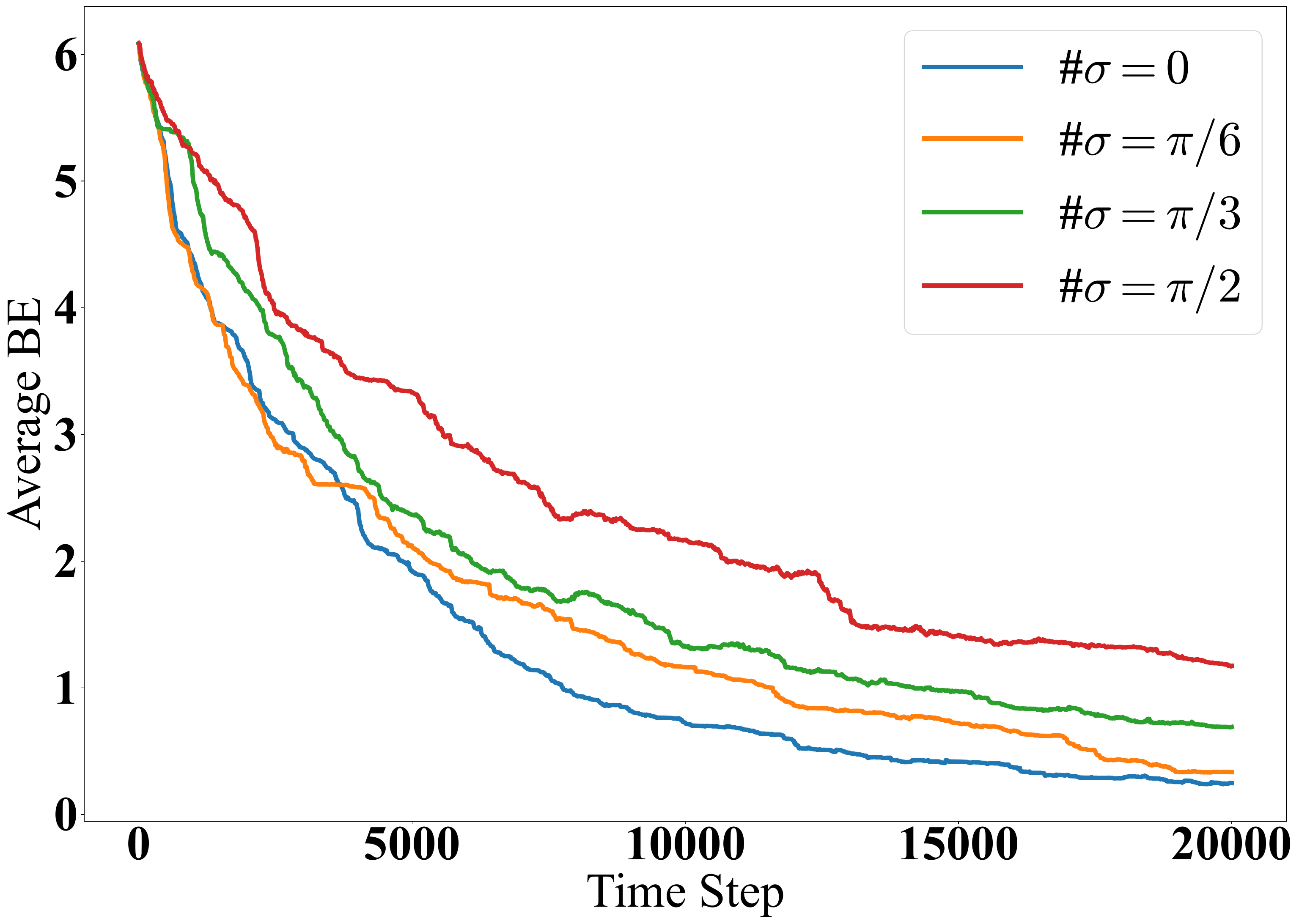}
    \caption{Average prediction error (left) and belief entropy (right) when the robot sensing is under noise of different levels.}
    \label{fig:Diff_noise}
\end{subfigure}
\hspace{5px}
\begin{subfigure}{0.29\linewidth}
    \centering
    \includegraphics[width=0.96\linewidth]{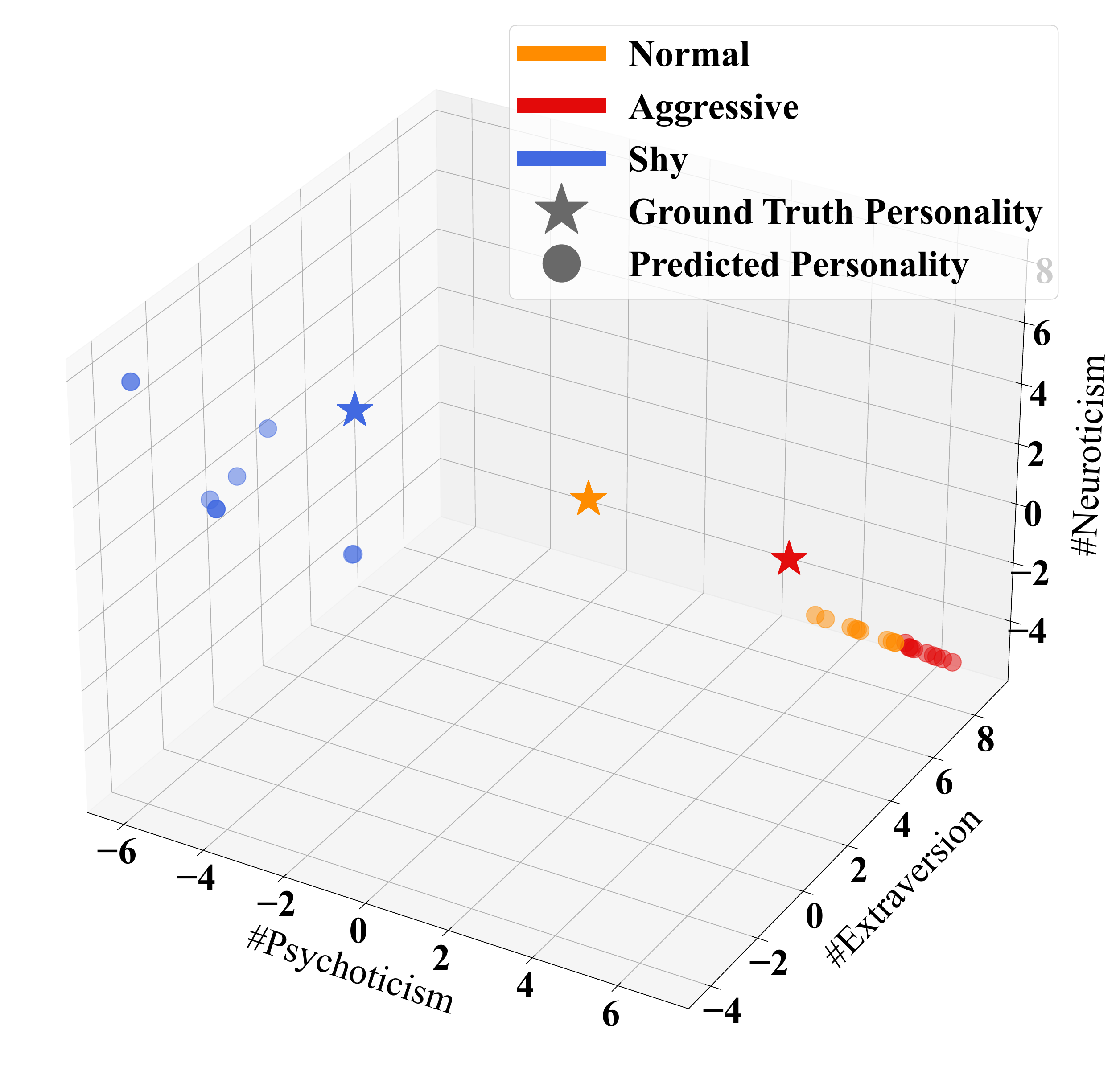}
    \caption{Prediction result when simulating with RVO.}
    \label{fig:RVO_sim_MPPI_infer}
\end{subfigure}
\caption{Performance of the model to noises of various levels and model inconsistency (non-learning-based RVO model).}
\vspace{-10px}
\end{figure*}

In practice, We inject noise into the learned navigation simulator. Such noise is injected only into the simulator, but not the trajectory estimation procedure. As a result, we could mimic the sensing error, which brings additional difficulties to the personality estimation. To explore the effect brought by the noise, we firstly set up $4$ noise levels by changing the value of the standard deviation: $\sigma=0, \pi/6, \pi/3, \pi/2$, where $\sigma=0$ means there is no noise, and when $\sigma>\pi/2$, the original behavior of the human will be difficult to identify. We then test the average prediction error of our method as illustrated in \prettyref{fig:Diff_noise} (left). While the convergence speed becomes slower with the increase of noise scale, our method provides prediction results with small gap, which indicates that our method is robust enough to noises of various levels. \revised{We further investigate the changes in belief entropy (BE) under these levels of noise. \prettyref{fig:Diff_noise} (right) shows that BE increases as noise levels rise. This indicates that, when the robot interacts with simple and unchanging human behavior, its prediction uncertainty decreases rapidly and converges to a low value. In contrast, when confronted with more complex and flexible human behaviors, the uncertainty remains relatively high. We further verify this finding through evaluations related to autism, which is discussed in later sections.}

Then we investigate the case where the human model used for trajectory prediction is inconsistent with the actual human behavior model. We replace the human model in the simulator with the true RVO~\cite{crowd_sim_psn_trait_theory} and still use the learned navigation simulator for prediction and MPPI for robot control. The experiment is carried out in ring-like scenario with 30 humans. The humans' personalities are set to 3 categories (normal, aggressive, shy) uniformly. \prettyref{fig:RVO_sim_MPPI_infer} illustrates that our method could still distinguish humans with different personalities, and there are visible boundaries between different human categories. However, humans with personality ``normal'' are inferred to be overly aggressive. We think this is the main impact of model inconsistency -- robot's decision boundary for human personalities becomes blurred, i.e., it becomes more difficult to distinguish two types of humans with insignificant behavioral differences. Despite of that, our method can still classify humans by their personalities, although the exact type of personality might be inaccurate.

\section{User Evaluation with Neurotypical Adults}
In the previous part, we have demonstrated that our methods can generate interactive robot behaviors that are more efficient and accurate in detecting the overall personality distribution of a crowd compared to the passive baseline in the simulator. In this section, we show the results of two follow-up user studies which evaluate our method's personality prediction ability for human participants and the effectiveness of active interaction. 

\subsection{\label{sec:user_psn_pred}User Personality Prediction}
A user study was performed to validate our method's performance in predicting human participant's personality, where we set up a 2-D game in the simulator in which participants could control one human's movement in a crowd, and we involved an active robot to estimate the participant's personality. We recruited 20 participants from a university campus with a female proportion of $40\%$ and an age range of $22\sim31$. The participants were asked to sign an informed consent form first and would get an equivalent coupon based on the measurement of 25.6 USD per hour after the experiment was completed. This user study took approximately 20 minutes for each participant.

\paragraph{Study Goals}
The main aim of this study is to explore our method's performance in human personality prediction. We further aim to investigate users' attitudes towards the robot's active interaction behavior.

\paragraph{Experimental Design}
This study consisted of a personality test section and two parts of games. In the first stage, we used a version of the Eysenck Personality Questionnaire, EPQ-RS \cite{eysenck1985revised}, which contains 48 yes or no questions and is aligned with the personality model we used, to test the participant's personality from the perspective of psychology, of which the result was treated as a trustworthy personality ground-truth. Afterwards, the participant will enter the game section.

The game was built in a 2-D simulator with several agents (representing humans) moving around in a square area similar to~\prettyref{fig:scenarios}. Agents were plotted as solid circles and could be divided into three categories: a user-controlled agent, an active robot, and several RVO agents with random parameters. The robot's parameters were adjusted slightly to reduce some interaction enthusiasm while maintaining the inference ability. The agent under the user's control was highlighted with red color and all other agents were plotted in brown including our active robot. The participant could use the arrow keys on the keyboard to control the movement of the highlighted agent. Participants were instructed to immerse himself/herself in the game context and imagine it as a real crowd scenario while without being informed of the possible presence of the robot. They needed to complete two parts of games (with a randomly disrupted order) for different research purposes. Before starting the game, The participants will first enter a 160-second trial phase to familiarize themselves with the operations. 

\textbf{Game part \Rmnum{1}: Moving as daily life.} This part of the game was intended to test the participant's original personality using our active robot. The participant was randomly given a target block representing an interested area which will be automatically changed when he/she reached. The participant's task was moving towards the generated target area while handling the interactions with others where both of them are equally important with no priority. The participant was encouraged to act as consistently as possible with his/her real daily life and the behavior was not restricted. To improve the participant's engagement level~\cite{pagnoni2002activity, treadway2009worth}, the participant was instructed that there is a reward for moving towards the target and collisions might be punished where the relative reward/punishment value is measured by the participants themselves. Note that as various psychological studies have shown that there exists a close relationship between personality and reward setting, participants can be more immersed in incorporating their own personalities into the game when given some reward and punishment settings~\cite{deyoung2010testing, gray1970psychophysiological}. This part of the game lasted for 160 seconds.

\textbf{Game part \Rmnum{2}: Probing one human's personality.} This part was intended to explore human participants' behaviors when they are trying to estimate another human's personality. The participant was asked to probe another highlighted agent's personality with physical moving without being given any target areas. The participant's behaviors were recorded for analysis. This part lasted for 50 seconds.  

At the end of the user study, the participant was asked whether he/she felt any other agents exhibiting unacceptable behavior during the whole game, which would be used to measure human attitudes towards active interactions.

\paragraph{Hypotheses} We investigated the following hypotheses:

$\mathcal{H}_0$: In game part \Rmnum{1}, the user personality predicted by the proposed method shows a correlation with the ground truth, which indicates that our method has effectiveness when applied on unknown and complex human personality distributions.

$\mathcal{H}_1$: In game part \Rmnum{2}, the user's personality probing behavior is similar to our active robot.

\paragraph{Result Analysis} 
To better demonstrate the personality distribution and prediction result, we conducted a classification based on the PEN personality values. As the values of `P' and  `E' show positive correlation and both of them show negative correlation with `N' in the used behavior model~\cite{crowd_sim_psn_trait_theory}, we divide the personality into two classes based on values in Eysenck’s three factors: One is Active/Aggressive (referred to as $c_\text{agg}$) where the mean value of `P' and `E' is higher than that of `N'; Another is Shy/Tense (referred to as $c_\text{shy}$) where `N' is relatively higher. The results of EPQ-RS show that the proportion of participants with $c_\text{agg}$ is $55.0\%$.

\begin{figure*}[ht]
\centering
\begin{subfigure}{0.46\linewidth}
    \centering
    \includegraphics[width=0.90\linewidth]{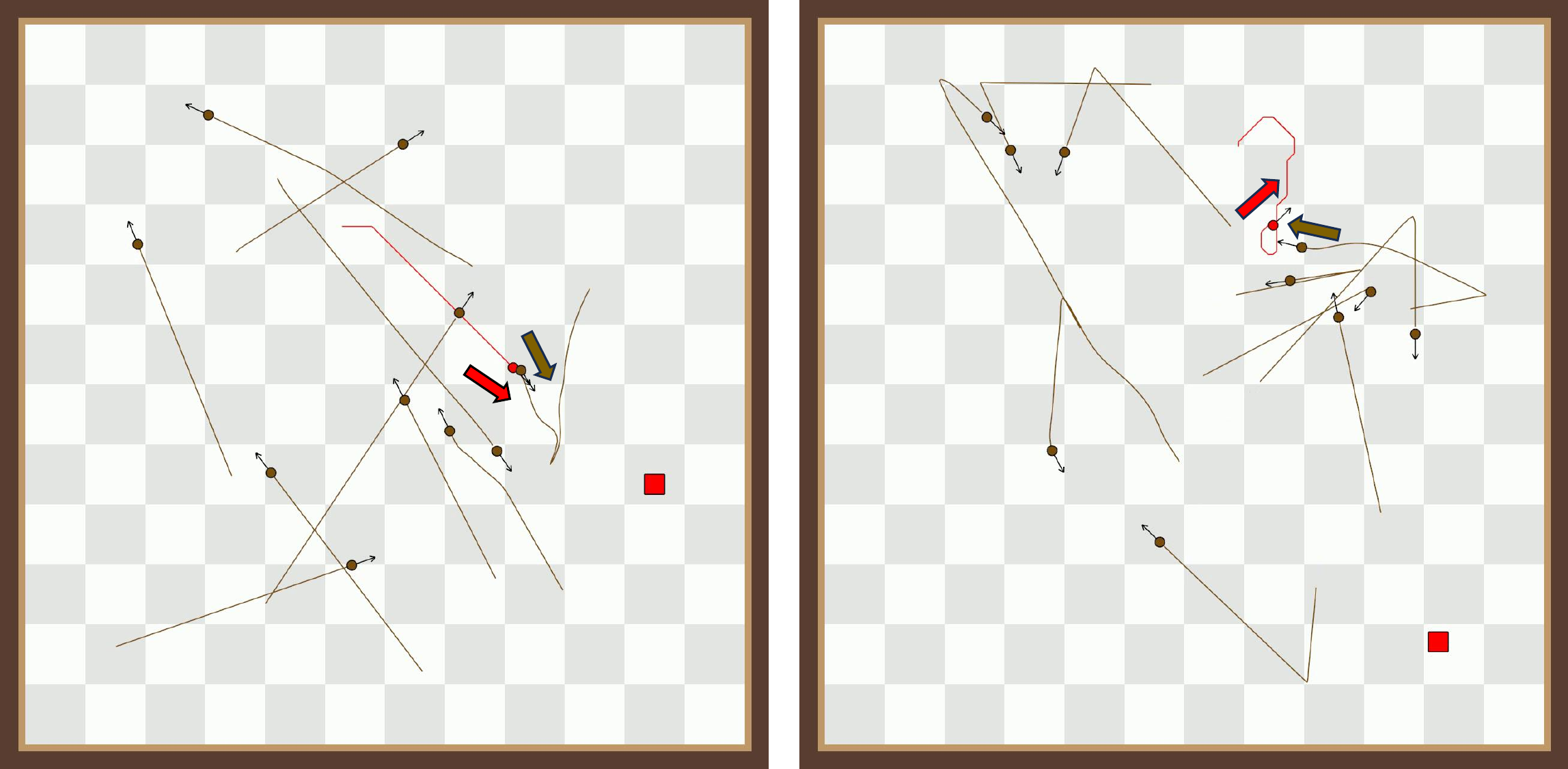}
    \caption{Behaviors of two participants in game part \Rmnum{1}. The red solid circle represents the agent which is under the user's control, where the red square is his/her interested area. The bold arrows show the local moving directions of the user (red) and his/her closest neighbor (brown). Left: A user disregards the collisions. Right: A user flees to the distance to avoid interactions.}
    \label{fig:user_trajs_daily_life}
\end{subfigure}
\hspace{5px}
\begin{subfigure}{0.50\linewidth}
    \centering
    \includegraphics[width=0.98\linewidth, trim=76 22 68 22,clip]{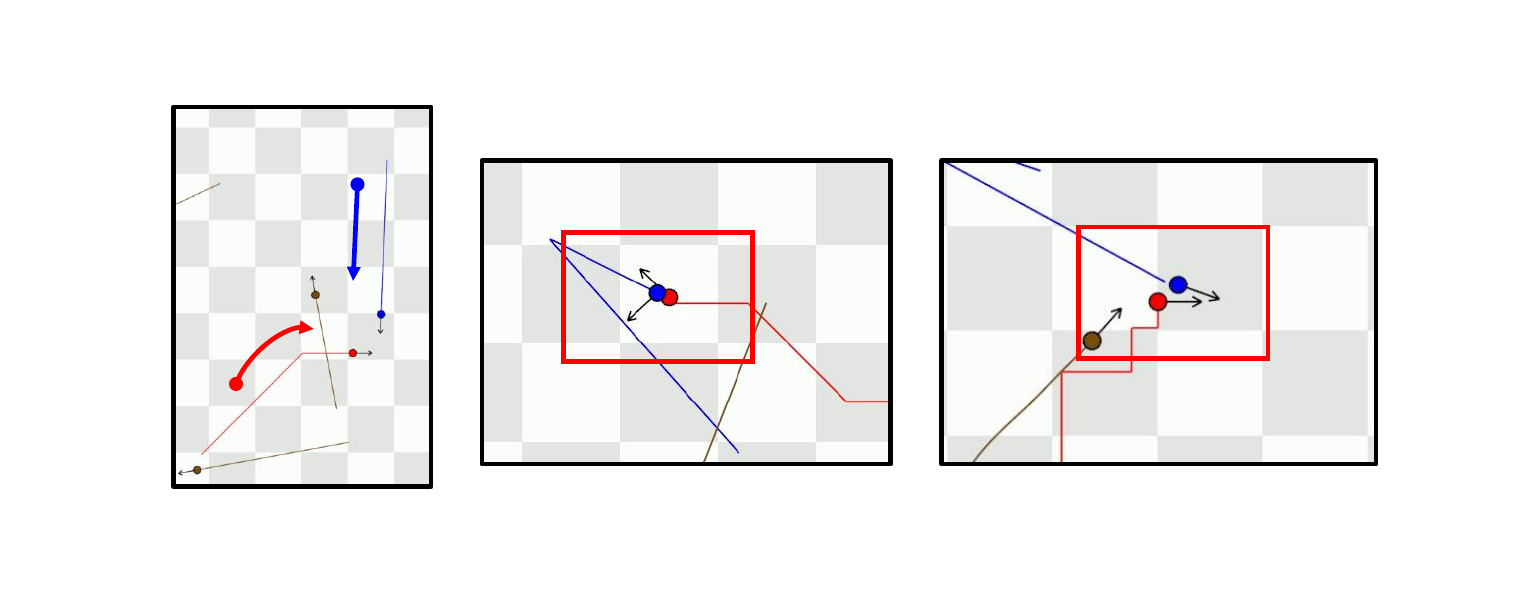}
    \caption{Trajectories of three participants when aiming to probe one human's personality. The red agent is controlled by a user and the blue one is the target agent that the user aims to probe. Left: An approaching behavior. Middle: A user collides with the target agent. Right: A user misses the opportunity to interact with the target agent.}
    \label{fig:user_probe_trajs}
\end{subfigure}
\caption{User behaviors observed in the study.}
\vspace{-10px}
\end{figure*}

\textbf{Result analysis for part \Rmnum{1}}. We first validate the correlation between the 3-D user personality values predicted by our active robot and ground truth. We calculate an overall Spearman correlation coefficient with associated \textit{p}-value, where the alternative hypothesis is set to be positive correlation. The result shows that the personality predictions have a moderate correlation with ground truth at a statistically significant rate ($rs=0.4654, p<0.001$), which supports the hypothesis $\mathcal{H}_0$. Then, we classify the personalities as mentioned before. The overall classification accuracy of the proposed method is $70.0\%$ and the weighted F1-score is 0.694. We observed rich user behaviors in this study, like emergency collision avoidance, avoiding the interaction with crowds in advance, or indifference to collisions. \prettyref{fig:user_trajs_daily_life} shows part of the participants' behaviors. The left figure demonstrates an aggressive behavior where the user disregards the collision with the neighbors and moves directly to the target; The right figure shows an avoiding moving pattern where the user moves away from the target area to avoid interactions. The above results indicate that our method can roughly infer a human participant's personality. 

\textbf{Result analysis for part \Rmnum{2}}. 
We recorded each participant's probing behavior and observed that all the participants demonstrated the behavior of approaching the target agent, which is similar to our active robot. The left figure in \prettyref{fig:user_probe_trajs} shows one participant's approaching behavior which aims to obtain more observations by influencing the target agent's movement. It shows that our active robot's behavior is intelligent and partially human-like, which supports $\mathcal{H}_1$. However, the participants' ability in active interaction is weaker than our active robot. On the one hand, user's performance in collision avoidance is less than satisfactory. As shown in \prettyref{fig:user_probe_trajs} (middle), the user collides with the target agent accidently when probing its personality. Our statistics show that $40\%$ of the participants caused more than 3 times of collisions during the probing process. On the other hand, user's inaccurate prediction of the target agent's future trajectory leads to a low interaction efficiency. \prettyref{fig:user_probe_trajs} (right) shows an example. Due to the user's inaccurate prediction of other agent's future trajectory, the user finds that the target agent has already moved to another location when he/she reaches the expected interaction position. By comparison, our active robot demonstrates robust collision avoidance and human behavior prediction abilities which have been discussed in \prettyref{sec:exp}.

Finally, we summarized the participants’ attitudes toward other agents' behaviors. $80\%$ of the participants stated that they did not experience unacceptable behavior during the whole gaming process. It proves that the proposed method can maintain human comfort while ensuring prediction accuracy.

\subsection{Behavior Perception Questionnaire}
To validate that the proposed active method brings more crucial information than the passive baseline, we conducted an additional user study where we generated some simulation videos with different settings (with active/passive robot) and asked users to label some humans' personalities from the video. The study was taken by the 20 participants who had been recruited in the previous game. The informed consent form was signed first and a coupon based on the measurement of 25.6 USD per hour was sent after the participant finishing the study. This user study was designed in the form of a questionnaire which took approximately 10 minutes to finish.

\paragraph{Study Goals}
This study aimed to compare the user's perception of personality from the given videos under two cases: scenarios with our active robot or with a passive robot. 

\paragraph{Experimental Design}
The basic task for the participant is using the given personality adjectives to classify the behavior of the specified agent in the crowd simulation. We generated crowd behaviors with different settings in simulator. The environment was set to be a square room. The number of agents had three categories: 5, 10, and 30, where 5 agents represented a sparse crowd which contains little natural interactions, and 30 agents is a dense case in which the interaction between agents was quite frequent. Agents that represent humans were controlled by RVO with two kinds of personality settings: active/aggressive, and shy/tense. We involved a robot in the crowd which is controlled either by our active method or passive baseline. For each category of crowd density and each robot type, we generate 6 cases randomly and get a video pool which contains totally 36 videos. All agents’ initial positions were randomly sampled and each human agent's parameter was sampled from the corresponding personality parameter distribution. 

In each video, all agents were plotted as solid circles. Two of the human agents were highlighted to attract user's attention while other agents looked the same. Participants were asked to view the generated videos and answer the corresponding questions. We sampled from the video pool randomly to select an instance and showed it to the participant. Each participant needed to watch 8 videos and the videos could be re-watched. In each instance, the participant was instructed to carefully observe the highlighted human's behavior and classify the highlighted human’s personality given the adjective candidates: \textit{Active/Aggressive}, \textit{Shy/Tense}, and \textit{Hard to decide}. The participant would further be asked the degree of difficulty he/she feel in evaluating each highlighted human's personality (ranged in 0-100), which could be understood as the evidence or confidence for supporting the evaluation. The participant was advised to be carefully observe the highlighted human’s trajectory and its interactions with the others, and mainly concern about its navigation and collision avoidance pattern.

\begin{figure*}[t]
\centering
\includegraphics[width=0.98\linewidth]{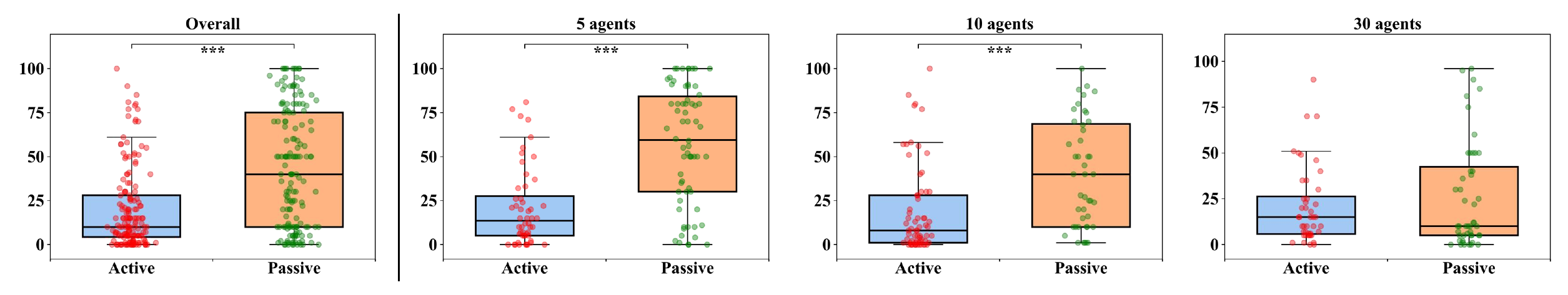}
\vspace{-10px}
\caption{\label{fig:boxplot_difficulty} Results of the evaluation difficulty in behavior perception questionnaire. The left figure shows the overall result, the right three figures show the results under different crowd densities. The vertical axis of each figure represents the evaluation difficulty score. Asterisks indicate statistical significant ($^{*}:p<0.05$, $^{**}:p<0.01$, $^{***}:p<0.001$).}
\vspace{-10px}
\end{figure*}

\paragraph{Hypotheses} 
We investigated following hypotheses:

$\mathcal{H}_2$: The participants show higher personality prediction accuracy and lower prediction difficulty in crowd simulations with our active robot than the passive robot.

$\mathcal{H}_3$: The participants show higher personality prediction accuracy and lower prediction difficulty in dense crowds than sparse crowds.

\paragraph{Result Analysis} 
Referring to conditions containing active and passive robot as $\mathcal{C}_\text{active}$ and $\mathcal{C}_\text{passive}$ respectively, we summarized the behavior perception results under $\mathcal{C}_\text{active}$ and $\mathcal{C}_\text{passive}$ separately in \prettyref{tab:questionnaire_results}. The \textit{Prediction Accuracy (PA)} was calculated to measure the proportion of agents whose personality is successfully predicted, while the \textit{Average Difficulty (AD)} was used to measure the averaged difficulty for participants to evaluated the agent's personality.

\textbf{Overall results.} The first column in \prettyref{tab:questionnaire_results} shows the overall results. Compared with the prediction accuracy under $\mathcal{C}_\text{passive}$ ($63.9\%$), the situation under $\mathcal{C}_\text{active}$ is obviously better ($81.5\%$). A Pearson’s chi-squared test is performed which indicates that there is a significant difference on the user prediction under the two conditions ($\chi^2=12.4557, p<0.001$). Meanwhile, the average evaluation difficulty under $\mathcal{C}_\text{active}$ is 20.1, which is less than half of the value under $\mathcal{C}_\text{passive}$ (42.9). We perform a one tailed Mann-Whitney U-test and it shows that our active method performs better than the passive baseline in evaluation difficulty at a statistically significant rate ($U=7776.0, p<0.001$). The evaluation difficulty values of each participant are visualized in \prettyref{fig:boxplot_difficulty} (left), which also demonstrates that the difficulty for user's evaluation is relatively lower under the condition of $\mathcal{C}_\text{active}$. The above results support our hypothesis $\mathcal{H}_2$.

\textbf{Results under different crowd densities.} We further break down the results to analyze the changing of evaluation results with different numbers of agents, where 5 agents represents a sparse crowd and 30 agents is a quite dense case. As shown in \prettyref{tab:questionnaire_results}, with the growing of agent numbers, both \textit{PA} and \textit{AD} under $\mathcal{C}_\text{active}$ have no obvious changes, while the indicators under $\mathcal{C}_\text{passive}$ show a significant improvement. The \textit{PA} and \textit{AD} under $\mathcal{C}_\text{active}$ significantly outperform the values under $\mathcal{C}_\text{passive}$ in the case of 5 agents, and the gap becomes smaller as the crowd becomes more dense. The statistical test results show a similar pattern. The evaluation difficulty values with different number of agents are shown in \prettyref{fig:boxplot_difficulty}. It also demonstrates that the level of evaluation difficulty under $\mathcal{C}_\text{active}$ stays stable with the changing of agent densities, while it shows a clear downward trend under $\mathcal{C}_\text{passive}$. The above results support our hypothesis $\mathcal{H}_3$, which is consistent with our previous claim in \prettyref{sec:psn_est}: Sparse environment contains less natural interaction between humans which requires more robot's active information gathering behaviors. 

\begin{table}[H]
\centering
\setlength{\tabcolsep}{4px}
\begin{tabular}{c|cc|cc|cc|cc}
\toprule
 \multirow{2}{*}{Conditions} & \multicolumn{2}{c|}{Overall} & \multicolumn{2}{c|}{5 agents} & \multicolumn{2}{c|}{10 agents} & \multicolumn{2}{c}{30 agents}\\
 {} & PA(\%)$^{***}$ & AD$^{***}$ & PA(\%)$^{***}$ & AD$^{***}$ & PA(\%) & AD$^{***}$ & PA(\%) & AD \\
\midrule
Active & 81.5 & 20.1 & 78.8 & 21.0 & 83.3 & 19.1 & 81.8 & 20.7 \\
Passive & 63.9 & 42.9 & 40.9 & 56.8 & 77.3 & 40.5 & 83.3 & 26.2 \\
\bottomrule
\end{tabular}
\caption{\label{tab:questionnaire_results}{The summary of results in the behavior perception study. Asterisks indicate statistical significant ($^{*}:p<0.05$, $^{**}:p<0.01$, $^{***}:p<0.001$).}}
\vspace{-15px}
\end{table}
\section{\revised{User Evaluation with Autistic and Neurotypical Participants}}
\revised{In the previous sections, we validated the effectiveness of our approach to the personality estimation problem through experiments conducted in both simulation and with typical adults. These experiments demonstrate the technical feasibility of our method for internal state estimation. As mentioned before, the potential application of our approach is in the diagnosis and treatment of autism, where we plan to deploy our idea on real robots with models built on interaction data collected from autistic participants in future work. The robot will actively interact with autistic participants through multiple modalities--visual, auditory, and tactile inputs--to predict their internal states. As a prior to this next step, in this work we conducted studies in autism using our current framework, where we fine-tune the algorithm's settings and apply a PC game-based user experiment with both autistic and neurotypical participants to explore meaningful preliminary results. Our studies were approved by Tsinghua University’s Institutional Review Board (IRB).}

\revised{A comparative user study was conducted with autistic and neurotypical participants to validate whether our method could identify the difference between the two groups based on their predicted internal states. We partially followed the game setup described in the previous section, where the participant could control one human’s movement in a crowd in the game, and we involved an active robot to estimate the participant’s internal state. Besides, the experiment's target and settings were adjusted to fit the specific needs of the groups in this study. We recruited 7 participants with autism, aged 9 to 20, from a special education school, 6 of whom were male. All participants with ASD included in this study have received a formal diagnosis from a licensed clinician. In addition, 8 neurotypical participants, aged 10 to 16, were recruited from regular high schools with 6 of them being female. Participants, or their family members/guardians, were asked to sign an informed consent form first and would get a participation fee based on the measurement of 14.1 USD per hour upon completing the experiment. The study was conducted offline, each participant was required to take part in the experimental PC game for three consecutive days, completing a total of six sessions, with two sessions per day.}

\paragraph{\revised{Study Goals}}
\revised{The primary goal of this study is to investigate whether our method can detect differences in the estimated internal states between autistic and neurotypical participants. If such differences are identified, we aim to further explore their meaning and significance.}

\paragraph{\revised{Experimental Design}}
\revised{The experiment consists of two parts: an intelligence test and a PC game. We first assessed the participants' intelligence quotients (IQs) using the Combined Raven's Test (CRT)~\cite{wang2007report}, a widely used non-verbal intelligence test that evaluates abstract reasoning and problem-solving abilities. The test consists of 72 questions and is suitable for a wide age range and individuals with varying cognitive abilities. Then, the participants were asked to play the game for 6 times in different time slots.}

\revised{The game framework is similar to the one described in \prettyref{sec:user_psn_pred}. It was built with a square area which contains a user-controlled agent (operated via keyboard), an active robot, and several RVO agents. The user-controlled agent was highlighted in red, while all other agents, including the active robot, were shown in brown. The participant’s task was using the arrow keys to control the highlighted agent, to move toward some randomly generated target areas iteratively while avoiding collisions with other agents. To accommodate the specific groups in this study, we made several modifications, including the following:}
\begin{enumerate}
    \item \revised{To ensure the participants' comfort during the interaction, we first adjusted the initial weights of the robot's optimization objectives as before. Additionally, we implemented a dynamic parameter adjustment procedure where the weight of ``active gathering'' was adjusted adaptively throughout the interaction to reduce the robot's enthusiasm. Specifically, we introduced two new parameters, $m_r$ and $\Tilde{m}$. The parameter $m_r$ measures the current adequacy of the robot's interaction, defined as follows:
    \begin{equation}
        m_r=
        \begin{cases}
            m_r+2, &\text{if $\|p_r-p_\text{user}\|<0.45d_H$} \\
            \max(0, m_r-0.3), &\text{otherwise}
        \end{cases}
        \notag
    \end{equation}
    where $\|p_r-p_\text{user}\|$ denotes the distance between the robot and the user. The threshold $\Tilde{m}$ represents a margin value for interaction adequacy. The weight of active information gathering, $w_\text{active}$, is adjusted adaptively based on $m_r$ and $\Tilde{m}$:
    \begin{equation}
        w_\text{active}=\max(0, 1.0-\frac{m_r}{\Tilde{m}})
        \notag
    \end{equation}
    This adjustment implies that as interactions between the robot and user increase, $w_\text{active}$ decreases rapidly, reducing the robot's activeness and shifting its focus more toward navigation. When the robot distances itself from the user, $m_r$ decreases slowly, allowing the robot's activeness to gradually return to its normal level. These adjustments prioritized general navigation over active interaction to create a more comfortable experience.}
    \item \revised{We modified the main objective and context of the game. Rather than asking participants to imagine themselves acting in the real world, we reframed the experiment as a ``pure game scenario''. Participants were instructed that it was a PC game in which they needed to reach the given target areas while avoiding collisions with others. To reinforce the game concept, we added collision effects and set up a scoring system: 30 points were deducted for collisions, and 100 points were awarded for reaching the target.}
    
    \revised{In this context, the estimated internal state no longer reflects the participant's ``real personality''. Instead, we refer to it as an ``internal state that influences the participant's behavior in the game context'', and we investigate the features of this internal state.}
\end{enumerate}

\revised{Both autistic and neurotypical participants were recruited. Before starting the experiment, autistic participants completed a trial phase lasting at least 30 minutes over 3 days to familiarize themselves with the game content and controls. Neurotypical participants were allowed unlimited familiarization time until they got satisfied, with the longest lasting no more than 5 minutes. This process ensured that all participants understood the game content, rules, and could operate effectively. Then, each participant played the PC game for a total of 6 sessions across 3 different time slots (2 sessions per slot), with each session lasting 6-7 minutes.}

\paragraph{\revised{Hypotheses}}
\revised{We investigated following hypotheses:}

\revised{$\mathcal{H}_4$: The estimated internal states of autistic and neurotypical participants exhibit different distributions.}

\revised{$\mathcal{H}_5$: The estimated internal states of autistic and neurotypical participants demonstrate different patterns both within a single game and across multiple game sessions.}

\paragraph{\revised{Result Analysis}}
\revised{Through the experiments, we obtained 6 pieces of data for each participant, including both their in-game behaviors and their estimated 3-D internal states. Based on this data, we conducted the following analyses:}

\begin{figure*}[ht]
\centering
\begin{subfigure}{0.60\linewidth}
    \centering
    \includegraphics[width=0.48\linewidth]{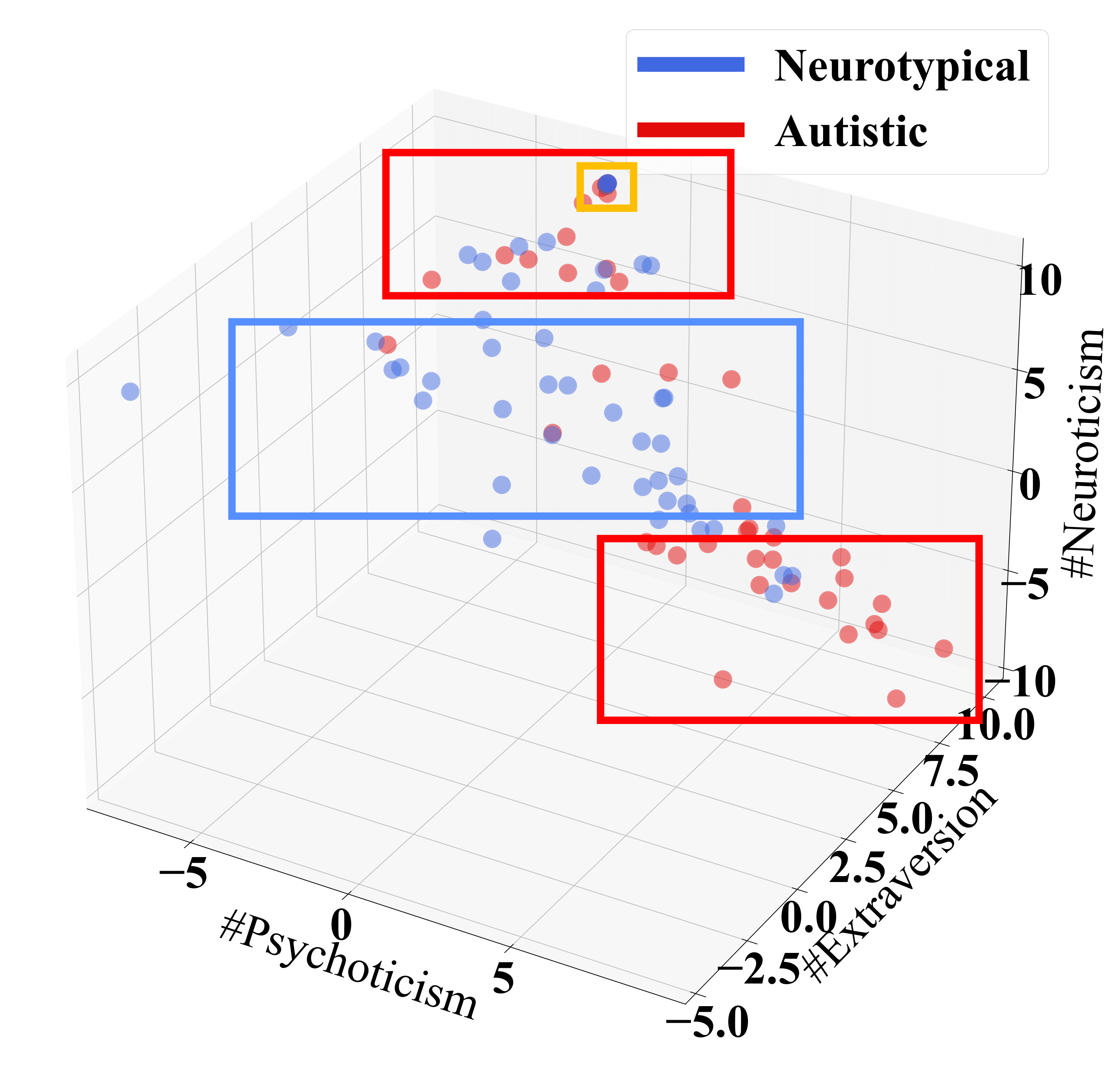}
    \includegraphics[width=0.48\linewidth]{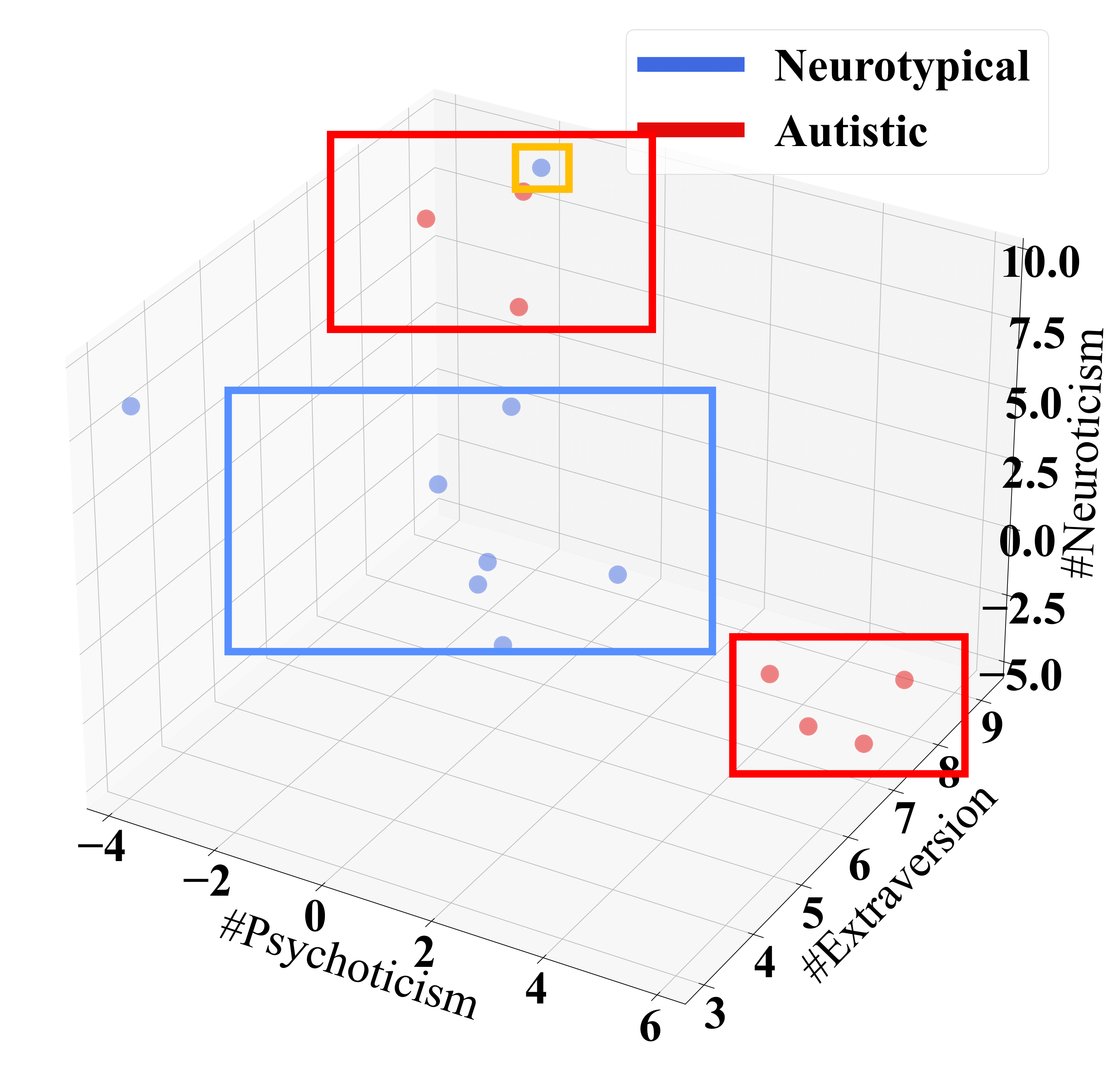}
    \caption{\revised{Overall (left) and average (right) internal state estimation for autistic (red) and neurotypical (blue) participants.}}
    \label{fig:asd_psn_3d}
\end{subfigure}
\hspace{5px}
\begin{subfigure}{0.36\linewidth}
    \centering
    \includegraphics[width=0.96\linewidth]{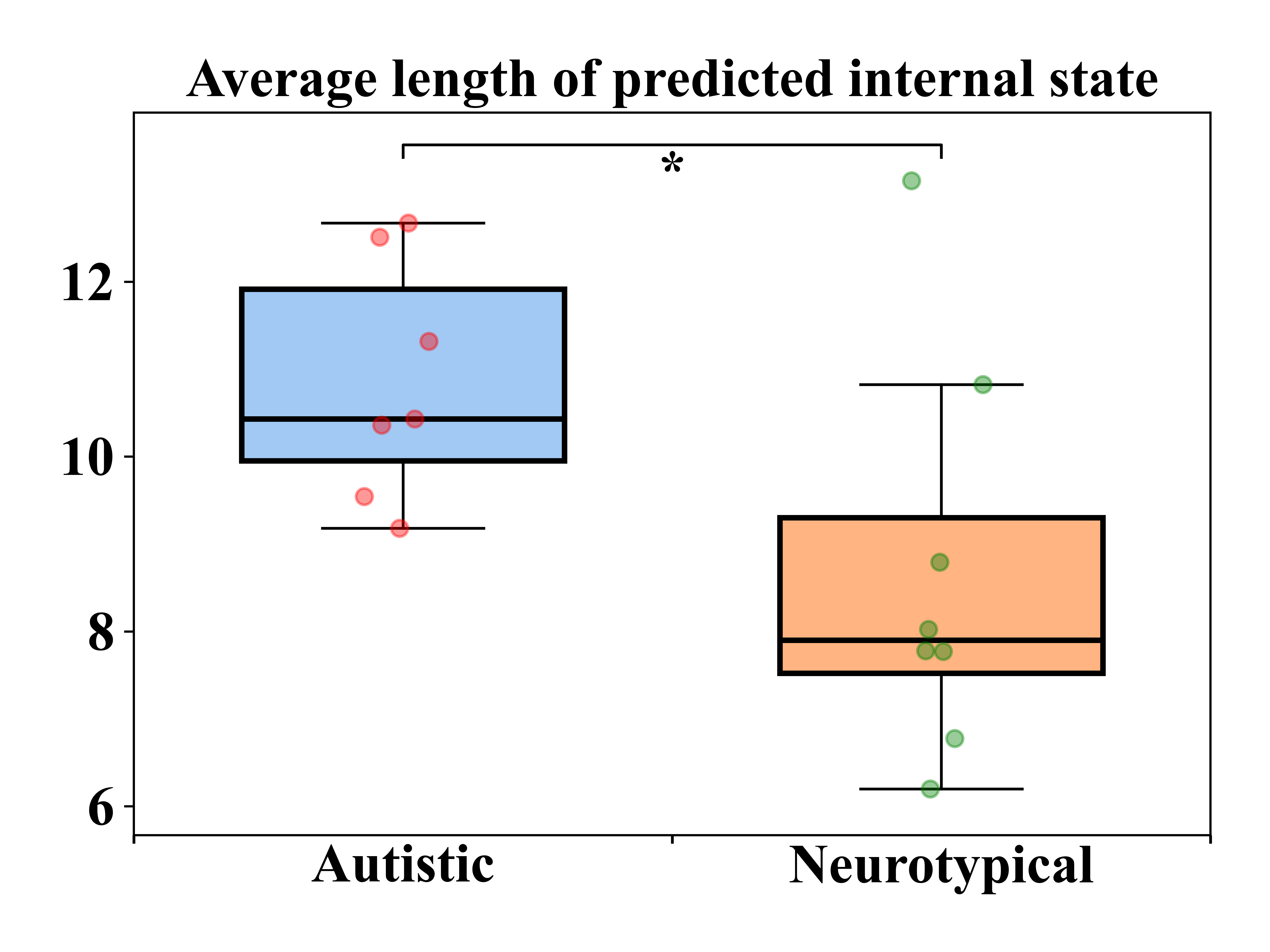}
    \caption{\revised{The average length of each participant's predicted internal states.}}
    \label{fig:boxplot_averlen}
\end{subfigure}
\caption{\revised{Visualization for the predicted internal states.}}
\vspace{-10px}
\end{figure*}

\revised{\textbf{Analysis of internal state distributions}. We plotted each participant's estimated internal states in 3-D space. \prettyref{fig:asd_psn_3d} (left) displays all session estimations for each participant, while \prettyref{fig:asd_psn_3d} (right) shows the average estimation for each participant. Red and blue dots represent autistic and neurotypical participants, respectively. We found that the estimations of neurotypical participants generally cluster near the origin point (within the blue frame), while autistic participants' estimations are closer to the boundary (within the red frame). This suggests that the robot's predictions for autistic participants are more extreme, with higher predicted values. To validate this, we calculated the average length of each participant's predicted internal states. The results, shown in \prettyref{fig:boxplot_averlen}, confirm that autistic participants have generally higher length values than neurotypical participants. A one-tailed Mann-Whitney U-test supports this trend at a statistically significant rate with $p<0.05$. Typically, simple and unchanging behavior patterns, such as consistently aggressive or tense behavior throughout the game, can drive the robot's predictions toward extreme values, as the robot's predictions will continuously shift in the same direction. We believe this result is mainly caused by the restricted and repetitive behavior patterns characteristic of autism. In the game, autistic participants tended to maintain a consistent movement pattern when facing various different situations due to these behaviors. For instance, some consistently rushed toward targets without regard for collisions, while others persistently avoided other agents even when the target was nearby. In contrast, neurotypical participants exhibited more complex and adaptive behaviors, adjusting flexibly to different situations. The above findings support our hypothesis $\mathcal{H}_4$.}

\revised{In addition, we observed one outlier: a neurotypical participant with extreme predicted values (within the yellow frame in \prettyref{fig:asd_psn_3d} and at the top right in \prettyref{fig:boxplot_averlen}). Notably, this participant had a considerably lower intelligence score than the other neurotypical participants (85 compared to an average of 122.8). This suggests that extreme predicted values in this experiment may correlate with intelligence quotients, as participants with autism often experience intellectual disabilities, generally scoring lower than neurotypical participants. This relationship will be explored further in our future work. Anonymized experimental information for participants is provided in \prettyref{tab:asd_infor}.}

\begin{table}[H]
\centering
\setlength{\tabcolsep}{6px}
\begin{tabular}{cccc|cccc}
\toprule
\multicolumn{4}{c|}{Autistic} & \multicolumn{4}{c}{Neurotypical} \\
id & gender & age (year) & intelligence & id & gender & age (year) & intelligence \\
\midrule
1 & M & 16 & 70 & 1 & F & 10 & 130 \\
2 & M & 20 & 60 & 2 & F & 15 & 120 \\
3 & F & 10 & 70 & 3 & F & 16 & 125 \\
4 & M & 20 & 60 & 4 & F & 15 & 130 \\
5 & M & 14 & 90 & 5 & F & 15 & 105 \\
6 & M & 12 & -- & 6 & M & 15 & 120 \\
7 & M & 9 & 85 & 7 & M & 15 & 85 \\
{} & {} & {} & {} & 8 & F & 15 & 130 \\
\bottomrule
\end{tabular}
\caption{\label{tab:asd_infor}\revised{Anonymized experimental information for participants. For gender, M: male, F: female.}}
\vspace{-15px}
\end{table}

\begin{figure*}[ht]
\centering
\includegraphics[width=0.38\linewidth]{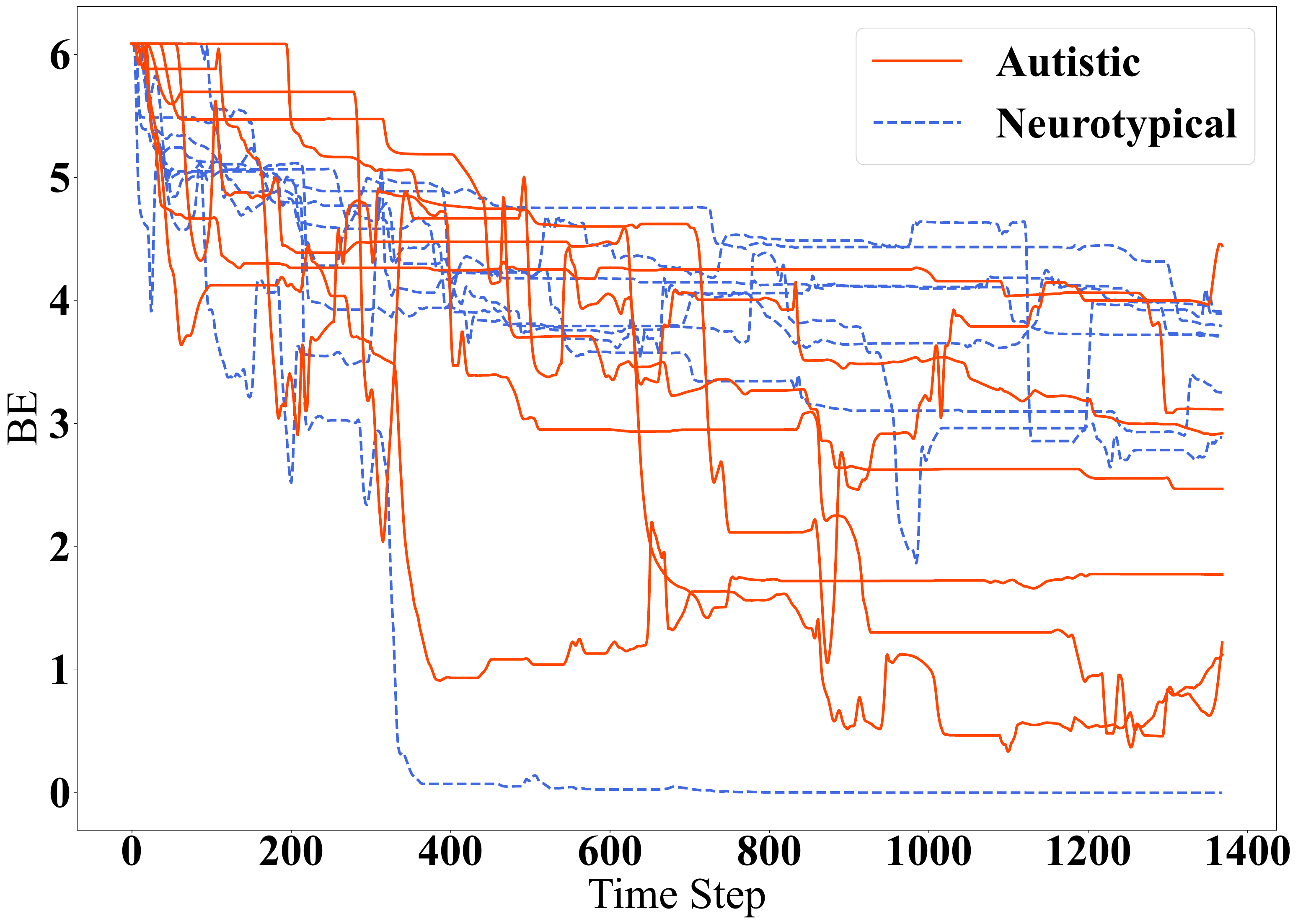}
\hspace{25px}
\includegraphics[width=0.38\linewidth]{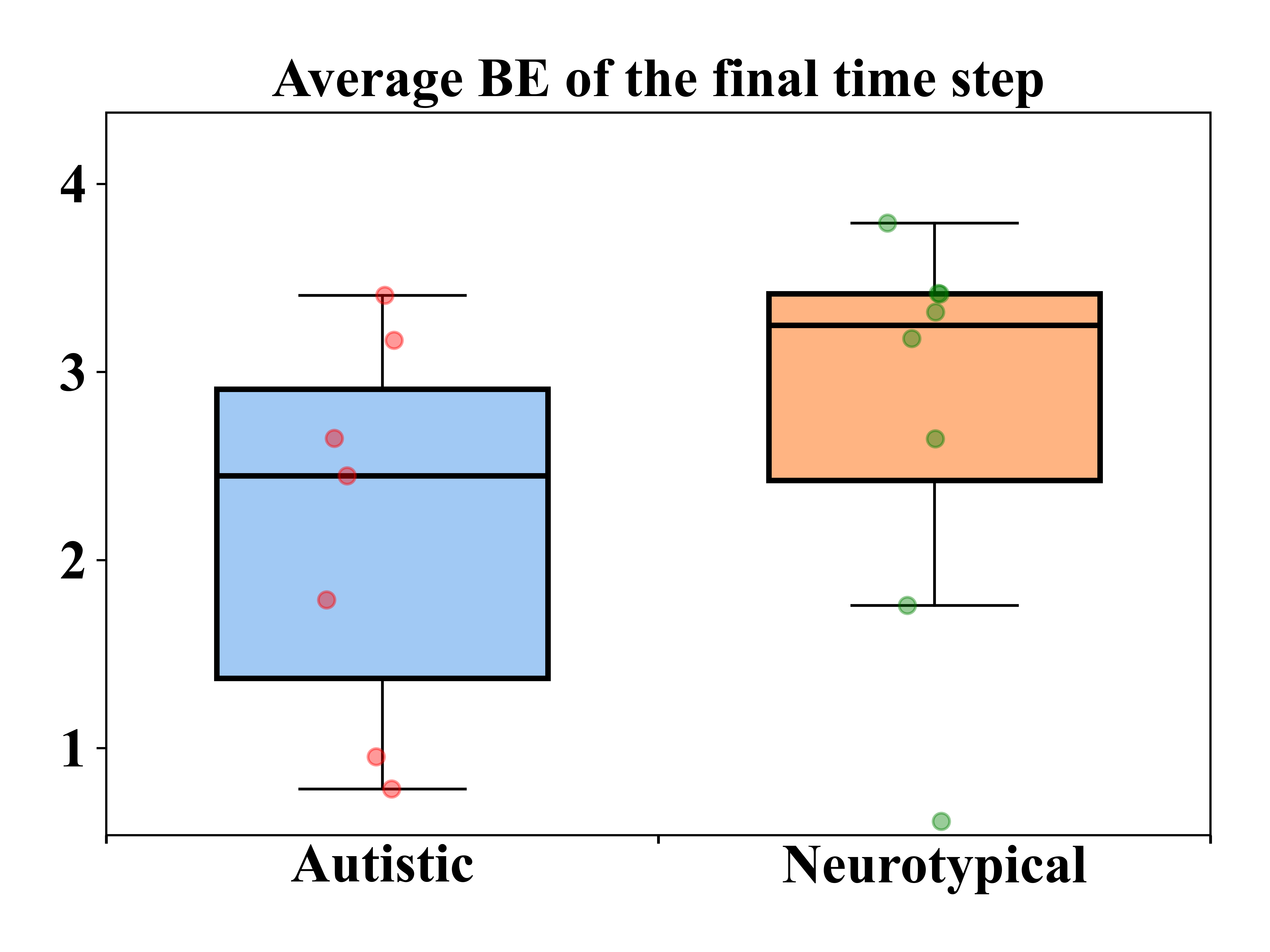}
\caption{\label{fig:asd_entropy}\revised{\textbf{Left}: The BE curve of each participant in one game session. \textbf{Right}: The average BE in the last game time step of each participant.}}
\vspace{-10px}
\end{figure*}

\revised{\textbf{Belief entropy change within single case}. We further investigated the behavioral characteristics of autistic and neurotypical participants within a single game session. As our active robot continuously updated its belief of each participant's internal state during the game, we selected one game session to plot each participant's belief entropy (BE) changes over time, which represent the robot's prediction uncertainty. As shown in \prettyref{fig:asd_entropy} (left), the BEs for autistic participants (red) show a greater reduction and converge to a lower value, while the curves for the neurotypical group (blue) are relatively higher. A similar trend was observed in a previous simulation experiment in \prettyref{sec:model_corr}, where we found that humans with higher behavioral noise exhibited slower decreases in prediction uncertainty, which then converged to a higher value. It indicates that the behavior of autistic participants tends to be more consistent and simple within a single game session, allowing the robot’s predictions to converge quickly. In contrast, neurotypical participants demonstrate more complex behaviors and adjust their behavior pattern according to different situations, which slows the decrease in prediction uncertainty. Note that the lowest blue curve in \prettyref{fig:asd_entropy} (left) represents the low-intelligence outlier case mentioned before. We further calculated the average BE at the final time step of each game for each participant, shown in \prettyref{fig:asd_entropy} (right), which confirms that the values for autistic participants are generally lower than those for neurotypical participants.}

\revised{\textbf{Prediction consistency across multiple games}. Except for single-game behavior, we also investigated the consistency of estimated internal states across multiple games for both autistic and neurotypical groups. With 6 estimations per participant, we used the intraclass correlation coeficient (ICC) to analyze consistency within each group separately. The results, shown in \prettyref{tab:asd_icc}, indicate that the autistic group exhibits higher consistency across games compared to the neurotypical group. This suggests that autistic participants are more likely to maintain consistent behavioral patterns over extended time frames and are less inclined to adapt their behavior to changing conditions. These findings further support that the characteristics observed in our predicted internal state distributions and trends may relate to the restricted and repetitive behavioral patterns typical of participants with autism. The above results support our hypothesis $\mathcal{H}_5$.}

\begin{table}[H]
\centering
\setlength{\tabcolsep}{6px}
\begin{tabular}{c|cccc}
\toprule
Group & ICC & F & pval & CI95\% \\
\midrule
Autistic & 0.827 & 29.649 & <0.001 & [0.71, 0.91] \\
Neurotypical & 0.561 & 8.674 & <0.001 & [0.39, 0.74] \\
\bottomrule
\end{tabular}
\caption{\label{tab:asd_icc}\revised{The ICC of both autistic and neurotypical groups.}}
\vspace{-15px}
\end{table}

\paragraph{\revised{Discussions}} \revised{During the experiment, we observed that participants' in-game behaviors often mirrored aspects of their real-life behavior. For instance, two autistic participants rarely demonstrated collision avoidance when encountering other agents in the game. Upon investigation and interviews, their guardians confirmed that these participants also rarely avoid people in real life. For example, if two people are walking and talking side by side on the road, they will directly pass between them. Additionally, one autistic participant consistently moved along the edges of the game interface--a behavior aligned with his tendency to walk along walls in real life. These observations suggest that our experiment successfully captured meaningful behavioral information. Furthermore, no autistic participants experienced emotional outbursts or impatience during the game, indicating that their experience was relatively comfortable and that the active robot did not cause significant emotional fluctuations.}

\revised{In summary, our PC-game-based evaluation provided valuable insights into user behaviors, identifying differences in the estimated internal states of autistic and neurotypical participants and highlighting their connection to the restricted and repetitive behavior patterns typical of autistic participants. These findings lay a foundation for our future work, which will involve using a real robot to interact with autistic participants.}
\section{Conclusion}
\revised{In this work, we introduce a scalable framework for active internal state estimation in human-robot interactions, focusing on personality estimation in group settings. Our approach uses the Eysenck 3-Factor model to link observed behaviors with personality traits, enabling robots to predict human behavior in real time and to gather information actively while considering human comfort. We demonstrate the efficacy of our method through evaluations on both typical and autistic groups. Specifically, our results show that active information gathering outperforms passive observation, reducing both personality prediction errors and uncertainty. The evaluations on typical adults demonstrate the method’s adaptability to diverse personality distributions. The preliminary study with autistic participants highlights its potential to identify behavior patterns associated with ASD, paving the way for applying real active robots in the diagnosis and treatment of ASD.}

\revised{Our work has several limitations. First, the performance of internal state estimation relies on the quality of the human behavior model. In our future work related to autism, we plan to establish a more realistic human behavior model using real-world interaction data. Second, we only use one internal state for estimation, while exploring the effect between multiple internal states is crucial in future work. Third, the robot's interactive behavior is limited to navigation policy in this work. We plan to apply multi-modal interaction strategies on the real robot in the future. Further, while our model accounts for human comfort, fine-tuning to achieve optimal interaction dynamics may require further adjustment across varied contexts. Finally, it would be interesting to explore multiple robots for more efficient cooperative internal state estimation in groups.}

\revised{We are excited to take a step toward active internal state estimation in group settings and to explore the scalable personality estimation, which lays the foundation for our future work in the robot-assisted diagnosis and treatment of ASD.}

\begin{acks}
This work was partially supported by the Natural Science Foundation of China (U2336214, 62332019).
\end{acks}

\bibliographystyle{ACM-Reference-Format}
\bibliography{references}


\end{document}